\RequirePackage{fix-cm}
\documentclass[smallextended]{svjour3}       
\smartqed  
\usepackage{graphicx}
\usepackage{amsmath}
\usepackage{dsfont}
\usepackage{algorithmic}
\usepackage{booktabs}
\usepackage{multirow}
\usepackage{mathrsfs}
\usepackage{amssymb}
\usepackage{color}
\usepackage{subfigure}
\usepackage[ruled]{algorithm2e}
\usepackage[colorlinks,linkcolor=blue]{hyperref}

\newcommand{\re}[1]{\textcolor{black}{#1}}
\newcommand{\newre}[1]{\textcolor{black}{#1}}

\usepackage{bbding}
\usepackage{amsmath}
\usepackage{amssymb}
\usepackage{bm}
\usepackage{booktabs}
\usepackage{multirow}
\usepackage{multicol}
\usepackage{diagbox}
\usepackage[normalem]{ulem}
\usepackage{graphicx}
\usepackage{arydshln}
\usepackage{booktabs}
\usepackage{listings}
\useunder{\uline}{\ul}{}
\makeatletter
  \newcommand\figcaption{\def\@captype{figure}\caption}
  \newcommand\tabcaption{\def\@captype{table}\caption}
\makeatother
\begin{document}

\title{DCP-NAS: Discrepant Child-Parent Neural Architecture Search for 1-bit CNNs
}

\titlerunning{DCP-NAS}        

\author{Yanjing Li${\dagger}$ \and Sheng Xu${\dagger}$  \and Xianbin Cao$^{*}$ \and Li'an Zhuo \and Baochang Zhang$^{*}$ \and Tian Wang \and Guodong Guo
}

\authorrunning{Y. Li, S. Xu, X. Cao and B. Zhang} 

\institute{Yanjing Li, Sheng Xu, Xianbin Cao, Li'an Zhuo  and Tian Wang \at
              Beihang University, Beijing \\
              \email{\{yanjingli, shengxu, xbcao, lianzhuo, bczhang, wangtian\}@buaa.edu.cn}\\
              Baochang Zhang\at
              Beihang University, Beijing; 
              Zhongguancun Laboratory, Beijing; 
              Nanyang Institute of Technology, Nanyang\\
              \email{bczhang@buaa.edu.cn}\\
            Guodong Guo \at
            UNIUBI Research, Universal Ubiquitous Co., Hangzhou, Zhejiang, CHINA.    \\
           \email{guodong.guo@mail.wvu.edu}
            \\${\dagger}$ Co-first authors with equal contribution.
            \\$^{*}$ Corresponding authors.
}

\date{Received: date / Accepted: date}

\maketitle

    \begin{abstract}
    Neural architecture search (NAS) proves to be among the effective approaches for many tasks by generating an application-adaptive neural architecture, which is still challenged by high computational cost and memory consumption. 
    At the same time, 1-bit convolutional neural networks (CNNs) with binary weights and activations show their potential for resource-limited embedded devices. 
    One natural approach is to use 1-bit CNNs to reduce the computation and memory cost of NAS by taking advantage of the strengths of each in a unified framework, while searching the 1-bit CNNs is more challenging due to the more complicated processes involved.
    In this paper, we introduce {\em Discrepant Child-Parent Neural Architecture Search (DCP-NAS)} to efficiently search 1-bit CNNs,  based on a new framework of searching the 1-bit model (Child) under the supervision of a \re{real-valued} model (Parent). 
    Particularly, we first utilize a Parent model to calculate a tangent direction, based on which the tangent propagation method is introduced to search the optimized 1-bit Child. 
    We further observe a coupling relationship between the weights and architecture parameters existing in such differentiable frameworks. To address the issue, we propose a decoupled optimization method to search an optimized architecture.
    Extensive experiments demonstrate that our DCP-NAS achieves much better results than prior arts on both CIFAR-10 and ImageNet datasets. In particular, the backbones achieved by our DCP-NAS achieve strong generalization performance on person re-identification and object detection.

    \end{abstract}

    \begin{figure}[t]
        \begin{minipage}[t]{0.5\textwidth}
        \hspace{-5mm}
    	\includegraphics[width=0.9\linewidth]{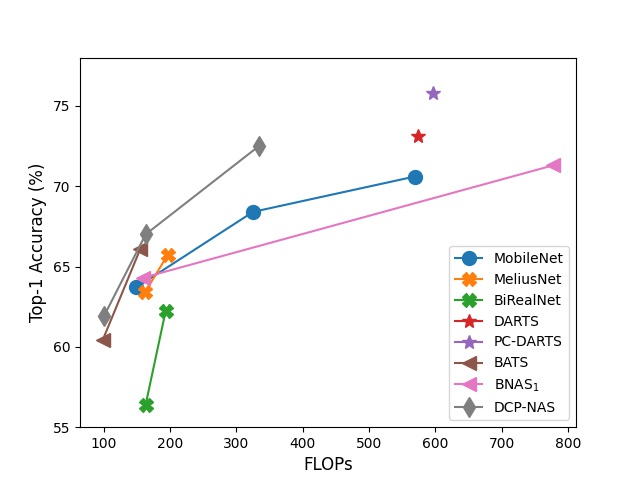}
    	\caption{Comparing DCP-NAS with \\lightweight CNNs and BNAS on Co-\\mputational cost vs. ImageNet Acc-\\uracy. DCP-NAS achieves the best \\performance-cost trade-off.}
    	\label{performance}
    	\end{minipage}
    	\hspace{-10mm}
    	\begin{minipage}[t]{0.6\textwidth}
    	\vspace{-35mm}
        \setlength{\tabcolsep}{1.1mm}{
    \begin{tabular}{ccccc}
    \toprule
    \begin{tabular}[c]{@{}c@{}}Network\\ Design\end{tabular} & Model      & 32-bit & 1-bit & Gap   \\ \hline
    \multirow{3}{*}{Hand-crafted}                            & ResNet-18  & 69.6   & 65.9  & -3.7  \\
                                                             & ResNet-34  & 73.3   & 69.0  & -4.3  \\
                                                             & ReActNet-A & 72.4   & 69.4  & -3.0  \\ \hline
    \multirow{3}{*}{Direct BNAS}                             & BNAS$_1$   & 67.1   & 64.3  & -2.9  \\
                                                             & BNAS$_2$   & 64.2   & 59.8  & -4.4 \\
                                                             &BNAS$_2$ v2   & 69.5   & 66.0  & -3.5\\
                                                            \hline
    \multirow{2}{*}{Auxiliary BNAS}                          & CP-NAS       & 68.0   & 66.5  & -1.5  \\
                                                             & \textbf{DCP-NAS}    & 68.5   & \textbf{72.4}  & \textbf{+4.0}  \\ \bottomrule
    \end{tabular}
    }
    \vspace{0.3mm}
    \centering
    \tabcaption{\re{Accuracy of both hand-crafted and search-based architectures with real-valued and their counterpart 1-bit CNNs on ImageNet.} \newre{Note that BNAS$_1$~\cite{chen2020BNAS,chen2020binarized}, BNAS$_2$~\cite{kim2020learning} and BNAS$_2$ v2~\cite{kim2021bnas} are different works.}}
    \label{inconsistency}
    \end{minipage}
    
    \end{figure}

    \section{Introduction}
    \label{intro}
    Neural architecture search (NAS) has attracted great attention with a remarkable performance in many computer vision tasks such as classification~\cite{zoph2016neural,tan2019efficientnet} and detection~\cite{chen2019detnas,wang2020fcos}. NAS attempts to design network architectures automatically and replace conventional hand-crafted counterparts at the expense of searching from a huge search space and a high computational cost in training. 
    Early efforts include sharing weights between searched and newly generated networks~\cite{cai2018efficient}. ProxylessNAS~\cite{cai2018proxylessnas} introduces a differentiable latency loss into NAS to search architectures on the target task instead of adopting the conventional proxy-based framework. PC-DARTS~\cite{xu2019pcdarts} takes advantage of partial channel connections to improve the memory efficiency by sampling a small part of the supernet to reduce the redundancy of the network search space. EfficientNet~\cite{tan2019efficientnet} introduces a new scaling method that uniformly scales all dimensions of depth, width and resolution using a simple yet highly effective compound coefficient to obtain efficient networks. 
    Besides these efficient NAS methods, developing effective NAS for efficient 1-bit convolutional neural networks (1-bit CNNs)~\cite{lin2017towards,xu2021layer,xu2022rbonn,xu2022ida,xu2023resilient} has drawn increasing attention, which retains the advantages of 1-bit CNNs on memory saving and computational cost reduction~\cite{chen2020BNAS,bulat2020bats}. 1-bit CNNs directly compress real-valued weights and activations (32-bit) of CNNs into a single bit and  directly decrease the memory consumption by $32\times$ and the computation cost by up to $58\times$~\cite{rastegari2016xnor}.

    \begin{figure}[t]
        \centering
        \includegraphics[scale=.4]{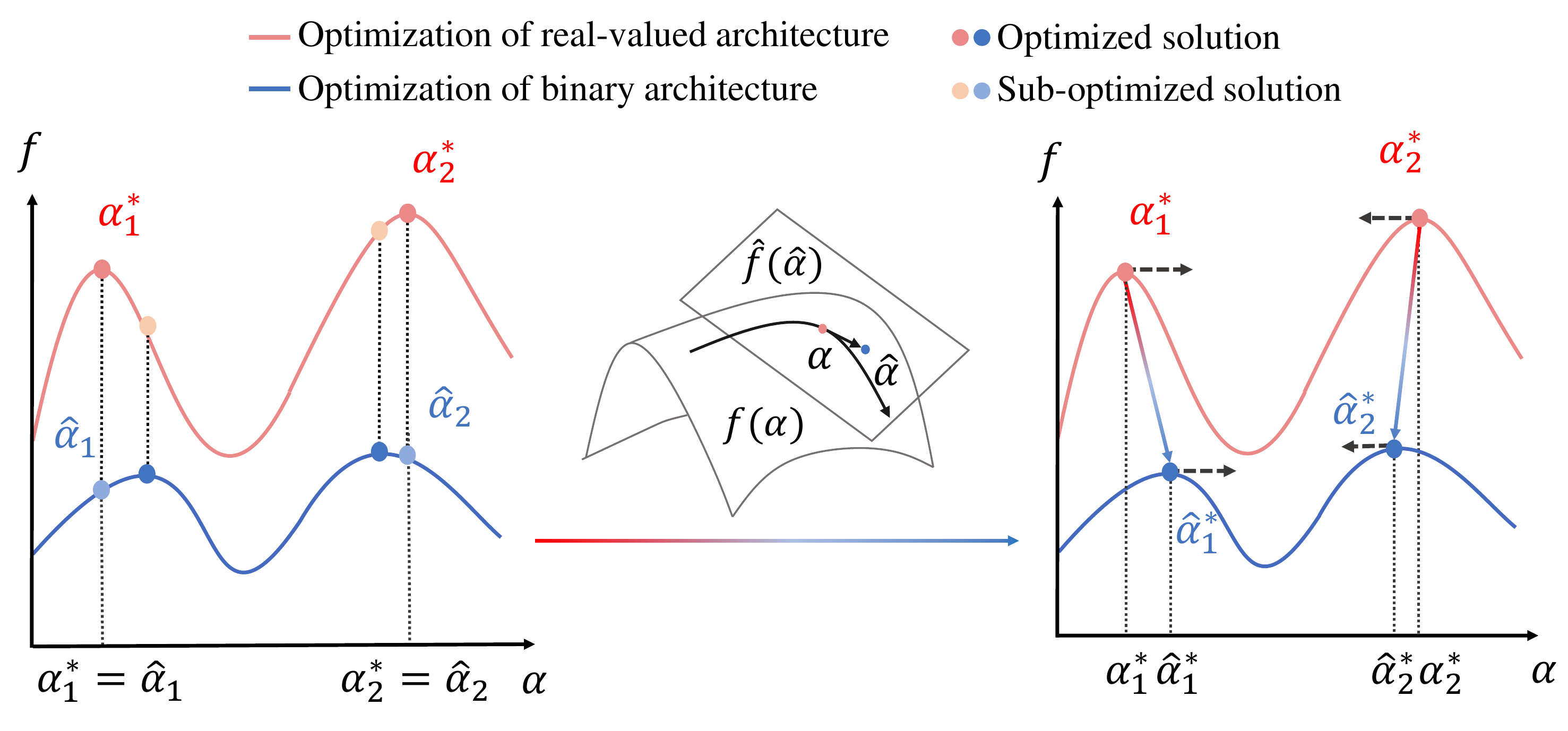}
        \caption{Motivation for DCP-NAS. We first show directly binarizing real-valued architecture to 1-bit is sub-optimal. Thus we use tangent propagation (middle) to find an optimized 1-bit neural architecture along the tangent direction, leading to a better-performed 1-bit neural architecture.} 
        \label{motivation}
    \end{figure}
    
    Comparatively speaking, 1-bit CNNs based on hand-crafted architectures have been extensively researched. 
    Binarized filters have been used in conventional CNNs to compress deep models~\cite{rastegari2016xnor,paper10,courbariaux2015binaryconnect,paper15}, which is widely considered as one of the most efficient ways to perform computing on embedded devices with low computational cost. 
    In~\cite{paper15}, the XNOR network is presented where both the weights and inputs attached to the convolution are approximated with binarized values. This results in an efficient implementation of convolutional operations by reconstructing the unbinarized filters with a single scaling factor. 
    In~\cite{gu2019projection}, a projection  convolutional neural network (PCNN) is proposed to implement \re{1-bit CNNs} based on a simple back propagation algorithm.~\cite{zhao2022towards} proposes Bayesian optimized 1-bit CNNs, taking the advantages of Bayesian learning to significantly improve the performance of extreme 1-bit CNNs. 
    \re{1-bit CNNs} show the advantages on computational cost reduction and memory saving, however, there still exists a significant gap (about 3\%$\sim$4\%) between 1-bit hand-crafted models and \re{real-valued} counterparts, as shown in Tab. \ref{inconsistency}.

    Binary neural architecture search (BNAS) is a simple and promising way to search the 1-bit network architectures. Previous BNAS methods are to directly search binary architecture~\cite{chen2020BNAS,kim2020learning}, {\em i.e.}, direct BNAS. BNAS$_1$~\cite{chen2020BNAS} attempts to search binary architectures from well-designed binary search spaces. Likewise, BNAS$_2$~\cite{kim2020learning} utilizes the diversity in the early search to learn better performing 1-bit neural architectures. However, such search strategy is only effective if an exhaustive search process is exploited, as revealed in~\cite{bulat2020bats}. Likewise, we observe a performance gap about 3\%$\sim$4\% compared with the real-valued counterparts from such direct BNAS, as depicted in Tab. \ref{inconsistency}. 
    
    \begin{figure}[t]
    \centering
    \includegraphics[scale=.35]{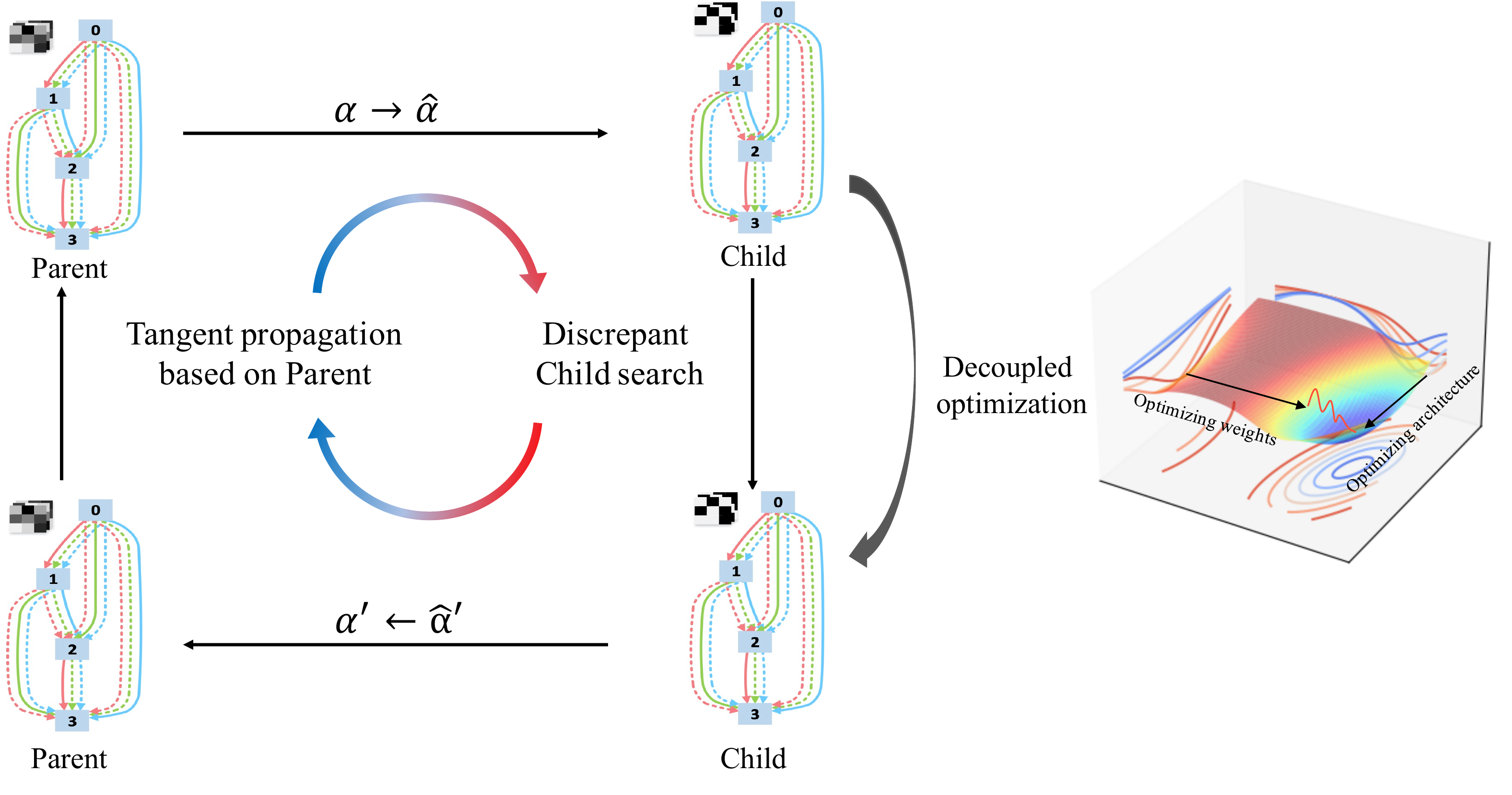}
    \hspace{8mm}
    \caption{The main framework of the proposed DCP-NAS, where $\alpha$ and $\hat{\alpha}$ denote real-valued and binary architecture, respectively. In one single round, we first conduct the real-valued NAS and generate the corresponding tangent direction. Then we learn a discrepant binary architecture via tangent propagation. In this process, real-valued and \re{1-bit CNNs} inherit architectures from their counterparts in turn.}
    \label{fig:dcpnas_framework}
    \end{figure}

    In this paper, we first introduce a Child-Parent framework to efficiently search a \re{1-bit CNN} in a unified framework.
    The real-valued models converge much faster than 1-bit models as revealed in~\cite{liu2021adam}, which motivate us to use the tangent direction of Parent supernet (real-valued model) as an indicator of the optimization direction for Child supernet (1-bit model).  

    We assume that all the possible 1-bit neural architectures can be learnt from the tangent space of the Parent model, based on which we introduce a {\em Discrepant Child-Parent Neural Architecture Search} (DCP-NAS) method to produce an optimized 1-bit CNN. Specifically, as shown in Fig.~\ref{motivation}, we utilize the Parent model to find a tangent direction to learn the 1-bit Child through tangent propagation, rather than directly binarizing the Parent to Child. Since the tangent direction is based on the second order information, we further accelerate the search process by Generalized Gauss-Newton matrix (GGN), leading to an efficient search process. 
    {Moreover, a coupling relationship between the weights and architecture parameters exists in such DARTS-based~\cite{liu2018darts} methods, leading to an asynchronous convergence and insufficient training process. To overcome this obstacle, we propose a decoupled optimization for training the Child-Parent model, leading to an effective and optimized search process. 
    The overall framework of our DCP-NAS is shown in Fig.~\ref{fig:dcpnas_framework}.}
    This paper is an extended version of our conference paper~\cite{zhuo2020cp}, where we make the following new contributions:

    1) We propose a new Discrepant Child-Parent model (DCP-NAS) to guide the binary architecture search, which further explore the discrepancy between real-valued and binary architectures.

    2) We utilize the Parent model to generate a tangent direction for Child, which learns the 1-bit neural architecture discrepancy through tangent propagation. The GGN method is further introduced to reduce the computation cost in the DCP-NAS optimization.

    3) We introduce a new decoupled optimization to address the asynchronous convergence in such differentiable NAS process, which further improve the performance of our DCP-NAS.

    4) We add a group of comprehensive experiments including new application on person re-identification and object detection tasks, which validate the acceleration, efficiency, and practicability of the proposed method.

    \section{Related Work}
    \label{sec:relatedwork}
    Extensive work has been reported on compressing and accelerating neural networks through quantization~\cite{rastegari2016xnor,li2022q,xu2023q,li2022q,xu2022rbonn,xu2022ida}, low-rank approximation~\cite{zhang2016accelerating,novikov2015tensorizing}, network pruning~\cite{PruningFilter,guo2016dynamic,han2015learning}, and more recently, architecture search~\cite{liu2018progressive,zoph2018learning}. Amongst them, quantization and architecture search  are most relevant to our work and are reviewed in this section.
    
    \subsection{\re{1-bit CNNs}} 

    BNN~\cite{paper10} first binarized the weights and activations of convolutional layers in CNNs. 
    BinaryNet based on BinaryConnect~\cite{courbariaux2015binaryconnect} is introduced to train DCNNs with binary weights, where activations are triggered at the run-time while parameters are computed at the training time~\cite{courbariaux2016binarized}. 
    In~\cite{bengio2013estimating}, the straight-through estimator (STE) is introduced to optimize BNNs. In the forward propagation, the sign function is directly applied to obtain binary parameters.
    In~\cite{rastegari2016xnor}, XNOR-Net is presented where both the weights and inputs of the convolution kernels are approximated with binary values, to improve the efficiency of the convolution operations. In stead of directly binarizing the weights into {$\pm 1$}, XNOR-Net reconstruct the binary weights with the mean absolute value (MAV) of each layer as the optimal value for $\alpha_l$.  
    XNOR++~\cite{bulat2019xnor} further designs an efficient estimator for layer-wise feature maps of the BNNs.
    Based on XNOR-Net, Bi-Real Net~\cite{liu2018bi} designs a new variant of residual structure to preserve the real activations before the sign function and a magnitude-aware gradient scheme \textit{w.r.t.} the weight to update the binarized weight parameters. 
    Instead of using a scale factor to estimate the real-valued kernel weights from the binary weights, HQRQ~\cite{li2017performance} adopts a high-order binarization method to achieve a more accurate approximation while retaining the advantage of the binaryized weights.
    Inspired by approximation of the real-valued weights helping the performance of BNNs, ABC-Net~\cite{lin2017towards} introduces multiple binary weights and activations for better estimation of the real-valued parameters to alleviate the degradation in prediction accuracy. 
   ~\cite{gu2019projection} introduces a quantization method based on a discrete back-propagation algorithm to learn a better BNNs.
    Furthermore, BONNs~\cite{zhao2022towards} extends a Bayesian method to the prior distributions of real-valued weights, features, and filters for constructing a BNNs in a comprehensive end-to-end manner, which further improves the performance. 
    ReActNet~\cite{liu2020reactnet} replaces the conventional PReLU~\cite{he2015delving} and the sign function of the BNNs with RPReLU and RSign with a learnable threshold, thus improving the performance of BNNs.
    LWS-Det~\cite{xu2021layer} effectively propose the angular and amplitude minimization technology to narrow the gap between real-valued kernels and 1-bit kernels.
   ~\cite{xu2021recu} explores the effect of dead weights, which refer to a group of weights, and proposes the rectified clamp unit (ReCU) to revive the dead weights for updating.

    \subsection{Neural Architecture Search} 
    Thanks to the rapid development of deep learning, significant gains in performance have been realized in a wide range of computer vision tasks, most of which are manually designed network architectures~\cite{krizhevsky2012imagenet,simonyan2014very,he2016deep,huang2017densely}. Recently, the new approach called neural architecture search (NAS) has been attracting increased attention. The goal is to find automatic ways of designing neural architectures to replace conventional hand-crafted ones. Existing NAS approaches need to explore a very large search space and can be roughly divided into three types of approaches: evolution based, reinforcement-learning-based and one-shot-based.
	
    In order to implement the architecture search within a short period of time, researchers try to reduce the cost of evaluating each searched candidate. Early efforts include sharing weights between searched and newly generated networks~\cite{cai2018efficient}. Later, this method was generalized into a more elegant framework named one-shot architecture search~\cite{brock2017smash,cai2018path,liu2018darts,pham2018efficient,xie2018snas,zheng2019dynamic}. In these approaches, an over-parameterized network or super network covering all candidate operations is trained only once, and the final architecture is obtained by sampling from this super network. For example,~\cite{brock2017smash} trained the over-parameterized network using a HyperNet~\cite{ha2016hypernetworks}, and~\cite{pham2018efficient} proposed to share parameters among Child models to avoid retraining each candidate from scratch. DARTS~\cite{liu2018darts} introduces a differentiable framework and thus combines the search and evaluation stages into one. Despite its simplicity, researchers have found some of its drawbacks and proposed a few improved approaches over DARTS~\cite{xie2018snas,chen2019progressive}. PDARTS~\cite{chen2019progressive} presents an efficient algorithm which allows the depth of searched architectures to grow gradually during the training procedure, with a significantly reduced search time. ProxylessNAS~\cite{cai2018proxylessnas} adopted the differentiable framework and proposed to search architectures on the target task instead of adopting the conventional proxy-based framework. 
    {IDARTS~\cite{xue2021idarts} focuses on decoupled optimization of NAS. In particular, IDARTS~\cite{xue2021idarts} decouple and backtrack the edge and operation parameters, in which the constraint $R(\cdot)$ of the edge parameter is utilized for controlling backtracking.}
	
    Binary neural architecture search replaces real-valued weights and activations with binarized ones, which consumes much less memory and computation resource to search \re{1-bit CNNs} and provides a more promising way to efficiently find network architectures. 
    These methods can be categorized into two classes, {\em direct binary architecture search} and {\em auxiliary binary architecture search}. Direct binary architecture search yields binary architectures directly from well-designed binary search spaces. 
    \newre{As one of the pioneer arts of this field, BNAS$_1$~\cite{chen2020BNAS}} effectively reduces the search time by channel sampling and search space pruning in the early training stages for a differentiable NAS. \newre{Another pioneer art BNAS$_2$~\cite{kim2020learning}} utilizes the diversity in the early search to learn better performing binary architectures. BMES~\cite{phan2020binarizing} learns an efficient binary MobileNet~\cite{howard2017mobilenets} architecture via the evolution-based search. However, the accuracy of direct binary architecture search can be improved by auxiliary binary architecture search~\cite{bulat2020bats}. BATS~\cite{bulat2020bats} designs a novel search space specially tailored to the \re{1-bit CNNs} and incorporates it into the DARTS framework. 
    \re{NASB~\cite{zhu2020nasb} explores an optimal architecture and connection for a convolutional group with introducing a three stage training scheme into BinaryNAS. 
    EBN~\cite{bulat2020high} propose Expert Binary Convolution to select data-specific expert binary and design a principled \re{1-bit CNNs} search mechanism that obtain a set of network architectures of preferred properties.
    BARS~\cite{zhao2020bars} design a joint search strategy for both macro-level and micro-level search space tailored for 1-bit CNNs and analyze the information bottlenecks relating to both the topology and layout architecture design choices.}
    
    Different from the aforementioned methods, our work is driven by the performance discrepancy between the 1-bit neural architecture and its real-valued counterpart. We introduce tangent propagation to explore the accuracy discrepancy, and further accelerate the search process by applying the GGN to the Hessian matrix in the optimization. Furthermore, we introduce a novel decoupled optimization to address the asynchronous convergence in such a differentiable NAS process, leading to better performed 1-bit CNNs. The overall framework leads to a novel and effective BNAS process.

    \section{Discrepant Child-Parent NAS}
    In this section, we first give the preliminaries of our method. Then, we describe our Child-Parent framework and the search space for binary NAS. Then we present the proposed architecture discrepancy-aware binary neural architecture search strategy together with tangent propagation. Finally, the decoupled optimization for solving the coupling relationship between weights and architecture parameters is introduced for effectively searching optimized binary neural architecture. For clarity, Tab.~\ref{tab:annotations} describes the main notations used in the following sections.

    \begin{figure*}[t]
	\centering
	\includegraphics[scale=.4]{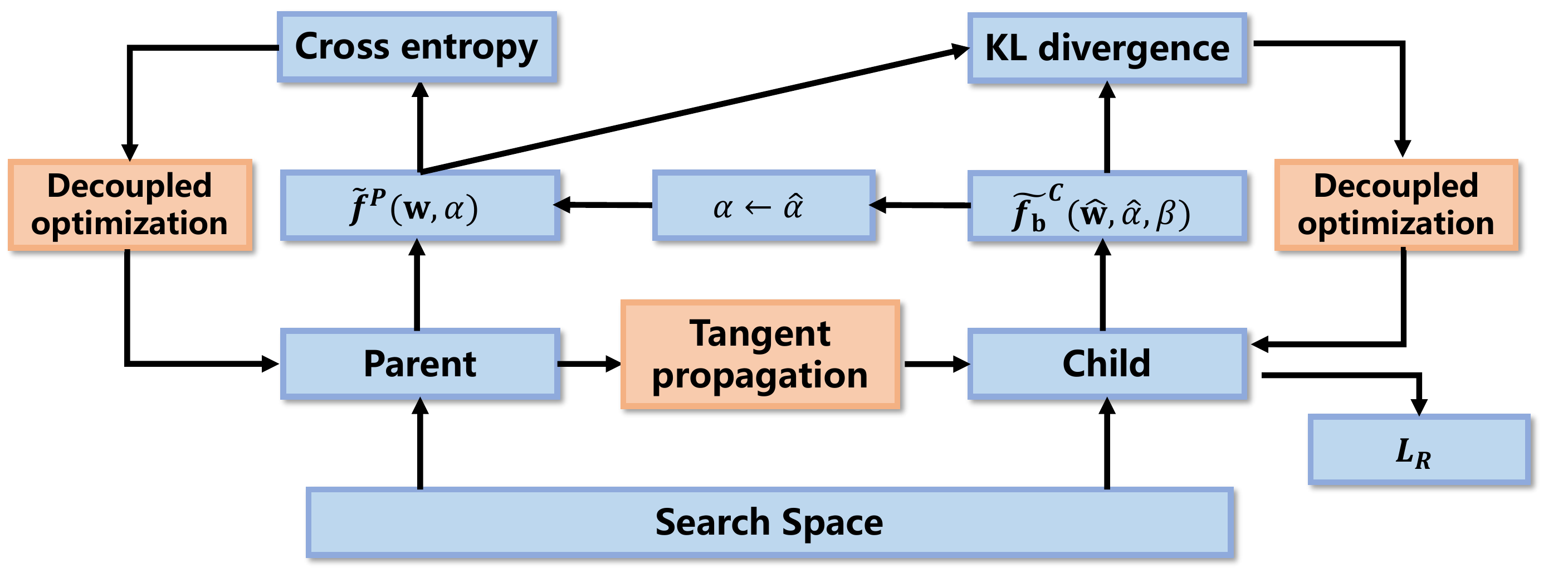}
	\caption{The main framework of Discrepant Child-Parent model. We show the key novelty of DCP-NAS, {\em i.e.}, tangent propagation and decoupled optimization, in orange.}
	\label{fig:dcp-model}
    \end{figure*} 

    \subsection{\re{Search Space}}
    \label{sec:3.3}
    We search for computation cells as the building blocks of the final architecture. Following~\cite{zoph2016neural,zoph2018learning,liu2018darts,real2019regularized}, we construct the network with a pre-defined number of cells and each cell is a fully-connected directed acyclic graph (DAG) $\mathcal{G}$ with $N$ nodes.  For simplicity, we assume that each cell only takes the outputs of the two previous cells as input and each input node has pre-defined convolutional operations for preprocessing. Each node $j$ is obtained by
	\begin{equation}\label{eq:node}
	\begin{aligned}
	    &{\bf a}^{(j)} = \sum_{i<j} o^{(i,j)}({\bf a}^{(i)})\\
	    &o^{(i,j)}({\bf a}^{i}) = {\bf w}^{(i,j)} \otimes {\bf a}^{i},
	\end{aligned}
	\end{equation}
    where $i$ is the dependent nodes of $j$ with the constraints $i<j$ to avoid cycles in a cell, and ${\bf a}^{j}$ is the output of the node $j$ node. ${\bf w}^{(i,j)}$ denotes the weights of the convolution operation between $i$-th and $j$-th nodes and $\otimes$ denotes the convolution operation. 
    Each node is a specific tensor like a feature map, and each directed edge $(i,j)$ denotes an operation $o^{(i,j)} (.)$, which is sampled from following $M = 8$ operations: 
    \begin{multicols}{2}
    	\small
    	\begin{itemize}
    		\item no connection (zero)
    		\item skip connection (identity)
    		\item $3\times3$ dilated convolution with rate $2$
    		\item $5\times5$ dilated convolution with rate $2$
    		\item $3\times3$ max pooling
    		\item $3\times3$ average pooling
    		\item $3\times3$ depth-wise separable convolution
    		\item $5\times5$ depth-wise separable convolution
    	\end{itemize}
    \end{multicols}
    We replace the depth wise separable convolution with a binarized form, {\em i.e.}, binary weights and activations. Note that, skip connection is identity mapping in NAS, instead of additional shortcut. The optimization of BNNs is more challenging than that of conventional CNNs~\cite{gu2019projection,rastegari2016xnor}, since binarization brings additional burdens to NAS. Following~\cite{liu2018darts},  to reduce the undesired fluctuation in the performance evaluation, we normalize the architecture parameter of $M$ operations for each edge to obtain the final architecture indicator as
    \begin{equation}\label{eq:soft_performance}
        \hat{o}^{(i,j)}_m({\bf a}^{(j)}) = \frac{exp\{{\alpha}_m^{(i,j)}\}}{\sum_{m'} exp\{{\alpha}_{m'}^{(i,j)}\}}o_m^{(i,j)}({\bf a}^{(j)}). 
    \end{equation}
    
    \subsection{\re{Preliminary}}
    \subsubsection{Neural architecture search (NAS)}
    
    \begin{table}[]
    \small
    \caption{A brief description of the main notations used in our paper. }
    \begin{tabular}{ll}
    \toprule
    ${\bf w}, \hat{\bf w}$: \re{real-valued} weight        
    & ${\bf a}, \hat{\bf a}$: \re{real-valued} activation        \\
    ${\bf b}^{\hat{\bf w}}$: binary weight 
    & ${\bf b}^{\hat{\bf a}}$: binary activation  \\
    $\beta$: scale factor                                               
    & \re{$\alpha, \hat{\alpha}$: real-valued and 1-bit architecture parameter} \\
    \re{$f(\cdot)$: real-valued NAS objective}                                       
    & \re{$\tilde{f(\cdot)}$: relaxed real-valued NAS objective} \\
    $f_b(\cdot)$: BNAS objective                                        
    & $\tilde{f}_b(\cdot)$: relaxed BNAS objective \\
    $p_n(\cdot), \hat{p}_n(\cdot)$: output logits       
    & $i, j$: node index \\
    $o^{(i, j)}(\cdot)$: operation                                      &$\mathcal{L}_{\operatorname{NAS}}$: searching objective of NAS   \\    
    $\mathcal{L}_{\operatorname{CP-NAS}}$: searching objective of CP-NAS
    &$\mathcal{L}_{\operatorname{DCP-NAS}}$: searching objective of DCP-NAS  \\
    $G(\cdot)$: objective of Child model training          
    &${\bf D}(\cdot)$: objective of tangent propagation \\                 
    $\mathcal{L}_R(\cdot)$: reconstruction error                  
    &${\bf H}(\cdot)$: Hessian matrix \\
    ${\bf I}(\cdot)$: Fisher information matrix 
    &\re{$R(\cdot)$: norm constraint in decoupled optimization}  \\ \hline
    \end{tabular}
    \label{tab:annotations}
    \end{table}
    
    Given a conventional CNN model, we denote ${\bf w}\in \mathcal{W} $ and $\mathcal{W}=\mathbb{R}_{C_{out}\times C_{in} \times K \times K}$ and ${\bf a}_{in}\in \mathbb{R}_{C_{in}\times W\times H}$ as its weights and feature maps in the specific layer. $C_{out}$ and $C_{in}$ represent the number of output and input channels of the specific layer. $(W, H)$ is the feature maps' width and height, and $K$ is the kernel size. We then have
    \begin{equation}\label{eq:conv}
        {\bf a}_{out} = {\bf a}_{in} \otimes {\bf w},
    \end{equation}
    where $\otimes$ is the convolution operation. We omit the batch normalization (BN) and activation layers for simplicity. 
    Based on this, a normal NAS problem is given as
    \begin{equation}\label{eq:nas}
        \max_{{\bf w}\in \mathcal{W}, \alpha \in \mathcal{A}} f({\bf w}, \alpha),
    \end{equation}
    where $f: \mathcal{W} \times \mathcal{A} \rightarrow \mathbb{R}$ is a differentiable objective function {\em w.r.t.} the network weight ${\bf w} \in \mathcal{W}$ and the architecture space $\mathcal{A} \in \mathbb{R}_{M \times E}$, where $E$ and $M$ denote the number of edges and operators, respectively. Considering that minimizing $f({\bf w}, \alpha)$ is a black-box optimization, we relax the objective function into $\tilde{f}({\bf w}, \alpha)$ as the objective of NAS
    \begin{equation}\label{eq:nas_objective}
    \begin{aligned}
    \min_{{\bf w}\in \mathcal{W}, \alpha \in \mathcal{A}} \mathcal{L}_{\operatorname{NAS}} &= - \tilde{f}({\bf w}, \alpha) \\ 
    &= - \sum^N_{n=1} p_{n}(\mathcal{X})\log(p_{n}({\bf w}, \alpha)),
    \end{aligned}
    \end{equation}
    where $N$ denotes the number of classes and $\mathcal{X}$ is the input data. $\tilde{f}({\bf w}, \alpha)$ represents the performance of a specific architecture with real-valued weights, where $p_{n}(\mathcal{X})$ and $p_n({\bf w}, \alpha)$ denote the true distribution and the distribution of network prediction, respectively.

    \subsubsection{Binary neural architecture search (BNAS)}
    The 1-bit model aims to quantize $\hat{{\bf w}}$ and $\hat{{\bf a}}_{in}$ into ${\bf b}^{\hat{\bf w}}\in \{-1,+1\}_{C_{out}\times C_{in} \times K \times K}$ and ${\bf b}^{\hat{{\bf a}}_{in}}\in \{-1,+1\}_{C_{in} \times H \times W}$ using the efficient XNOR and Bit-count  operations to replace \re{real-valued} operations. 
    \re{More specifically, all 1-bit CNNs are trained using the fake quantization strategy, {\em i.e.}, the weights and activations are represented by $\{-1,+1\}$ in FP32 format using sign$(\cdot)$ during training, as mentioned in~\cite{gholami2021survey}. Efficient XNOR and Bit-count operations are used in hardware deployment.}
    Following~\cite{courbariaux2015binaryconnect,courbariaux2016binarized}, the forward process of the 1-bit CNN is
    \begin{equation}\label{eq:bconv}
        {\hat{{\bf a}}_{out}} = \beta \circ {{\bf b}^{\hat{{\bf a}}_{in}}} \odot {{\bf b}^{\hat{\bf w}}},
    \end{equation}
    where $\odot$ is the XNOR, and bit-count operations and $\circ$ denotes the channel-wise multiplication. $\beta=[\beta_1,\cdots,\beta_{C_{out}}]\in \mathbb{R}^+_{C_{out}}$ is the vector consisting of channel-wise scale factors. ${\bf b} = {\rm sign(\cdot)}$ denotes the binarized variable using the sign function, which returns 1 if the input is greater than zero, and -1 otherwise. It then enters several non-linear layers, {\em e.g.}, BN layer, non-linear activation layer, and max-pooling layer. We omit these for simplification. Then, the output $\hat{{\bf a}}_{out}$ is binarized to ${\bf b}^{\hat{{\bf a}}_{out}}$ via the sign function. The fundamental objective of BNNs is calculating $\hat{\bf w}$. We want it to be as close as possible before and after binarization, such that the binarization effect is minimized. Then, we define the reconstruction error following~\cite{gu2019projection} as 
    \begin{equation}\label{eq:l_R}
        \mathcal{L}_R(\hat{\bf w},{\beta}) = \|\hat{\bf w}-\beta\circ {\bf b}^{\hat{\bf w}}\|^2_2.
    \end{equation}
    Based on the derivation above, the vanilla direct BNAS~\cite{chen2020BNAS,kim2020learning} can be defined as
    \begin{equation}\label{eq:bnas}
        \max_{\hat{{\bf w}}\in \mathcal{W}, \hat{\alpha} \in \mathcal{A}, \beta\in\mathbb{R}^+} f_{\bf b}({\hat{{\bf w}}}, \hat{\alpha}, \beta),
    \end{equation}
    where ${\bf b}^{\hat{{\bf w}}} = {\rm sign}(\hat{{\bf w}})$ is used for inference and $\hat{\alpha}$ is neural architecture with binary weights. Prior direct BNAS~\cite{chen2020BNAS} learning the BNAS from such objective as
    \begin{equation}\label{eq:bnas_objective}
        \max_{\hat{{\bf w}}\in \mathcal{W}, \hat{\alpha} \in \mathcal{A}, \beta\in\mathbb{R}^+} \tilde{f}_{\bf b}({\hat{{\bf w}}}, \hat{\alpha}, \beta) = \sum^N_{n=1} \hat{p}_{n}(\hat{\bf w}, \hat{\alpha}, \beta)\log(\hat{p}_{n}(\mathcal{X})),
    \end{equation}
    where we use similar denotations as Eq.~\ref{eq:nas_objective}. Eq.~\ref{eq:bnas_objective} means that vanilla direct BNAS only focus on the binary search space under the supervision of cross entropy loss, which is less effective due to the search process is not exhaustive~\cite{bulat2020bats}.

    \begin{figure}[t]
	\centering
	\includegraphics[scale=.40]{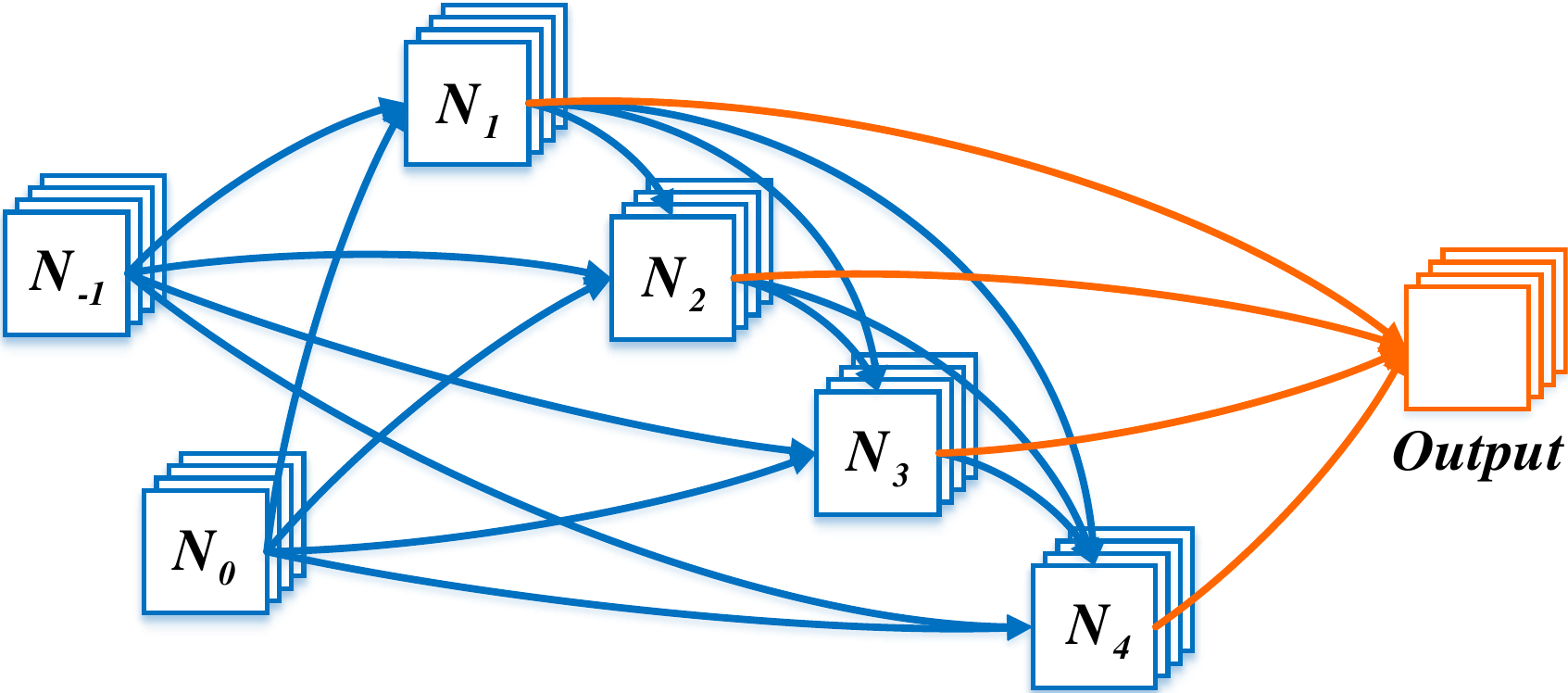}
	\caption{The cell architecture for DCP-NAS. One cell includes $2$ input nodes, $4$ intermediate nodes and $14$ edges (blue).}
	\label{fig:cell}
    \end{figure}  
	
    \subsection{Child-Parent Framework for Network Binarization}
    Network binarization, which calculates neural networks with binary weights and activations to fit the \re{real-valued} network, can significantly compress the CNNs. Prior work~\cite{zhao2022towards} usually investigates the binarization problem by exploring the \re{real-valued} model to guide the optimization of \re{1-bit CNNs}. Based on the investigation, we reformulate NAS-based network binarization as a Child-Parent model as shown in Fig. \ref{fig:dcp-model}. The \re{1-bit CNNs} and the \re{real-valued} counterpart are the Child and Parent model, respectively.

    Conventional NAS is inefficient  due to the complicated reward computation in network training where the evaluation of a structure is usually done after the network training converges. There are also some methods to perform the evaluation of a cell during the training of the network. 
   ~\cite{zheng2019multinomial} points out that the best choice in early stages is not necessarily the final optimal one, however, the worst operation in the early stages usually has a bad performance in the end. This phenomenon will become more and more significant as the training goes. Based on this observation, we propose a simple yet effective operation removing process, which is the key task of the proposed CP-model. \re{In a searching loop, we first sample the operation without replacement for each edge from the search space, and then train the sampled model generated by the corresponding supernet. Second, we validate our model on the validation set. Until all the operations are selected, we remove the operation in each edge with the worst performance, {\em i.e.}, in terms of the accuracy of both models on the validation dataset.}
    
    Intuitively, the representation difference between the Children and Parent, and how much Children can independently handle their problems, are two main aspects that should be considered to define a reasonable performance evaluation measure. Based on these analysis, we introduce the Child-Parent framework for binary NAS, which defines the objective as 
    \begin{equation}
        \begin{aligned}
        \hat{\bf w}^*, \hat{\alpha}^*, \beta^* &= \mathop{\rm arg min}_{\hat{{\bf w}}\in \hat{\mathcal{W}}, {\alpha} \in \mathcal{A}, \beta\in\mathbb{R}^+} \; \mathcal{L}_{\operatorname{CP-NAS}}(\tilde{f}^P({\bf w}, \alpha),\;\; \tilde{f}^C_{\bf b}(\hat{\bf w}, \hat{\alpha}, \beta)) \\
        &= \mathop{\rm arg min}_{\hat{{\bf w}}\in \hat{\mathcal{W}}, {\alpha} \in \mathcal{A}, \beta\in\mathbb{R}^+} \tilde{f}^P({\bf w}, \alpha) - \tilde{f}^C_{\bf b}(\hat{\bf w}, \hat{\alpha}, \beta),
        \end{aligned}
    \label{eq5}
    \end{equation}
    where $\tilde{f}^P({\bf w}, \alpha)$ denotes the performance of  the real-valued Parent model as pre-defined in Eq.~\ref{eq:nas_objective}. $\tilde{f}_{\bf b}^C$ is further defined as $\tilde{f}^C_{\bf b}({\hat{{\bf w}}}, {\alpha}, \beta) = \sum^N_{n=1} \hat{p}_{n}(\hat{\bf w}, {\alpha}, \beta)\log$ $(\hat{p}_{n}(\mathcal{X}))$ following Eq.~\ref{eq:bnas_objective}. 
    As shown in Eq.~\ref{eq5}, we propose the $\mathcal{L}$ for estimating the performance of the candidate architectures with binary weights and activations, which take both the real-valued architectures and \re{1-bit} architectures into consideration.

    \subsection{Tangent propagation for DCP-NAS}
    \label{sec3.4}
    In this section, we first propose the generation of the tangent direction based on the Parent model, and then present the tangent propagation for effectively searching the optimized architecture in the binary NAS. As shown in Fig. \ref{fig:dcpnas_framework}, the DCP-NAS novelty introduce tangent propagation and decoupled optimization, thus leading to a effective discrepancy-based search framework.
	The main motivation of DCP-NAS is to ``fine-tune" the Child model architecture based on the real-valued Parent rather than directly binarizing the Parent. Thus we first take advantage of the Parent model to generate the tangent direction from the architecture gradient of model as
    \begin{equation}
        \frac{\partial \tilde{f}^C({\bf w}, \alpha)}{\partial \alpha} = \sum_{n=1}^N \frac{\partial p_n({\bf w}, \alpha)}{\partial \alpha},
        \label{eq4}
    \end{equation}
    where $\tilde{f}({\bf w}, \alpha)$ is pre-defined in Eq. \ref{eq:nas_objective}.
	
    Then we conduct the second step, {\em i.e.}, tangent propagation for Child model. 
    For each epoch of binary NAS in our DCP-NAS, we inherit weights from the real-valued architecture $\hat{\alpha} \leftarrow \alpha$ and enforce the \re{1-bit CNNs} to learn similar distributions as real-valued networks
    \begin{equation}
    \begin{aligned}
        \max_{\hat{{\bf w}}\in \mathcal{W}, \hat{\alpha} \in \mathcal{A}, \beta\in\mathbb{R}^+} \;\;G(\hat{\bf w}, \hat{\alpha}, \beta) &= \tilde{f}^P({\bf w}, \alpha) log \frac{\tilde{f}^P({\bf w}, \alpha)}{\tilde{f}^C_{\bf b}(\hat{\bf w}, \hat{\alpha}, \beta)} \\
        &= \sum_{n=1}^N p_{n}({\bf w}, \alpha)\log(\frac{\hat{p}_{n}({\hat{{\bf w}}}, \hat{\alpha},\beta)}{p_{n}({\bf w},\alpha)}),
    \end{aligned}\label{eq6}
    \end{equation}
    where the KL divergence is employed to supervise the binary search process. $G({\hat{{\bf w}}}, \hat \alpha ,\beta)$ calculates the output logits similarity between real-valued network $p(\cdot)$ and \re{1-bit CNNs} $\hat{p}(\cdot)$, where the output of teacher is already given. 

    To further optimize the binary architecture, we constrain the gradient of binary NAS using the tangent direction as
    \begin{equation}
        \begin{aligned}
            \min_{\hat{\alpha} \in \mathcal{A}} {\bf D}(\hat{\alpha}) &= \|\frac{\partial \tilde{f}^P({\bf w}, \alpha)}{\partial \alpha} - \frac{\partial G({\hat{\bf w}}, \hat{\alpha},\beta)}{\partial \hat{\alpha}}\|_2^2. \\
    \end{aligned}
    \label{eq7}
    \end{equation}
    
    We use Eqs.~\ref{eq6} - \ref{eq7} to jointly learn the DCP-NAS and rewrite the objective function in Eq.~\ref{eq5} as
    \begin{equation}
    \begin{aligned}
        \mathcal{L}_{\operatorname{DCP-NAS}}&(\tilde{f}^P({\bf w}, \alpha),\;\; \tilde{f}^C_{\bf b}(\hat{\bf w}, \hat{\alpha}, \beta)) \\
        &= -G(\hat{\bf w},\hat{\alpha},\beta) + \lambda {\bf D}(\hat{\alpha}) + \mu \mathcal{L}_R({\hat{\bf w}, \beta}).
    \end{aligned}
    \label{eq9}
    \end{equation}
    Then we optimize the binary architecture $\hat{\alpha}$ along the tangent direction of the real-valued model, which inherits from the real-valued one. Note that, when we set $\lambda=0$, the Eq. \ref{eq9} is equivalent to the objective of original CP-NAS~\cite{zhuo2020cp}.
    As revealed in~\cite{liu2021adam}, the real-valued weights generally converge faster than the binary ones. Motivated by this observation, the tangent direction of Parent supernet can be used to approximate the optimization direction of the more slowly converged Child supernet. To conclude, in Eq. \ref{eq7}, we improve the optimization of Child architecture based on tangent direction of the Parent architecture, which lead Child supernet to be more efficiently trained.
    
    Considering the binary weights are learned by KL divergence, we optimize our DCP-NAS as
    \begin{equation}
    \begin{aligned}
        \nabla_{\hat{\alpha}}&\mathcal{L}_{\operatorname{DCP-NAS}}(\tilde{f}^P({\bf w}, \alpha),\;\; \tilde{f}^C_{\bf b}(\hat{\bf w}, \hat{\alpha}, \beta)) \\
        &=  -\frac{\partial G(\hat{\bf w},\hat{\alpha},\beta)}{\partial \hat{\alpha}} + \lambda\frac{\partial {{\bf D} (\hat{\alpha})}}{\partial \hat{\alpha}}\\
        &= - \frac{\partial G(\hat{\bf w},\hat{\alpha},\beta)}{\partial \hat{\alpha}} + 2\lambda(\frac{\partial G({\hat{\bf w}}, \hat{\alpha},\beta)}{\partial \hat{\alpha}} - \frac{\partial \tilde{f}^P({\bf w}, \alpha)}{\partial \alpha})\frac{\partial^2 G({\hat{\bf w}}, \hat{\alpha},\beta)}{\partial \hat{\alpha^2}}\\
        &= - \frac{\partial G(\hat{\bf w},\hat{\alpha},\beta)}{\partial \hat{\alpha}} + 2\lambda(\frac{\partial G({\hat{\bf w}}, \hat{\alpha},\beta)}{\partial \hat{\alpha}} - \frac{\partial \tilde{f}^P({\bf w}, \alpha)}{\partial \alpha}){\bf H}_G(\hat{\alpha}),
    \end{aligned}
    \label{eq10}
    \end{equation}

    \begin{equation}
        \nabla_{{\hat{{\bf w}}}}\mathcal{L}_{\operatorname{DCP-NAS}}(\tilde{f}^P({\bf w}, \alpha),\;\; \tilde{f}^C_{\bf b}(\hat{\bf w}), \hat{\alpha},\beta) = - \frac{\partial G(\hat{{\bf w}}, \hat{\alpha},\beta)}{\partial {\bf b}^{\hat{\bf w}}}\frac{{\partial {\bf b}^{\hat{\bf w}}}}{\partial \hat{\bf w}},\\
    \label{eq11}
    \end{equation}
    where
    \begin{equation}
    \begin{aligned}
    \small
        \frac{{\partial {\bf b}^{\hat{\bf w}}}}{\partial \hat{\bf w}} = {\bf 1}_{|\hat{\bf w}|\leq 1}.
    \label{eq12}
    \end{aligned}
    \end{equation}
    and $\lambda$ is a hyper-parameter, ${\bf H}_{\tilde{f_{\bf b}}}(\hat{\alpha}) = \frac{\partial^2 \tilde{f_{\bf b}}({\hat{\bf w}}, \hat{\alpha})}{\partial \hat{\alpha^2}}$ denotes the Hessian matrix. The process of DCP-NAS is outlined in Fig.~\ref{fig:dcpnas_framework}. 
    
    We minimize the difference of the gradient (tangent direction) $ \frac{\partial G({\hat{\bf w}}, \hat{\alpha},\beta)}{\partial \hat{\alpha}}$ and the gradient (tangent direction) $\frac{\partial \tilde{f}^P({\bf w}, \alpha)}{\partial \alpha}$, aiming to search the architectures (both real-valued NAS and binary NAS) in the same direction to generate a better 1-bit architecture. Note that $\alpha$ inherits from $\hat{\alpha}$ at the beginning of each real-valued NAS iteration, indicating that we only utilize w and $\alpha$ of real-valued NAS for heuristic optimization direction for 1-bit NAS instead of searching a better architecture for real-valued networks. Since a better tangent direction $\frac{\partial \tilde{f}^P({\bf w}, \alpha)}{\partial \alpha}$ is achieved, DCP-NAS can have a more suitable $\hat{\alpha}$ for \re{1-bit CNNs}. We note that $\alpha$ is different from $\hat{\alpha}$, which is not an optimized architecture for real-valued weights but an optimized architecture for binary weights.

    The expression above contains an expensive matrix gradient computation in its second term. Thus we introduce a first-order approximation of the Hessian matrix for accelerating the searching efficiency in Section~\ref{sec:pca}.

    \subsection{Generalized Gauss-Newton matrix (GGN) for Hessian Matrix}\label{sec:pca}
    
    Since the Hessian matrix is computationally expensive, this section mainly tries to accelerate the aforementioned Hessian matrix calculation by deriving a second-order expansion based on Eq.~\ref{eq11}. 
    
    Below we proved that the Hessian matrix of loss function is directly related to the expectation of the covariance of the gradient. Considering the loss function as the negative logarithm of the likelihood, let $\mathcal{X}$ be a set of input data of network and $p(\mathcal{X}; \hat{\bf w}, \hat{\alpha})$ be the predicted distribution over $\mathcal{X}$ under the parameters of the network are $\hat{{\bf w}}$ and $\hat{\alpha}$, {\em i.e.,} output logits of the head layer.

    With omitting $\hat{{\bf w}}$ for simplicity, the Fisher information of the probability distribution set $P = \{p_n(\mathcal{X};\hat{\alpha}), n \in N\}$ can be described by a matrix whose value at the $i$-th row and $j$-th column
    \begin{equation}
        I_{i,j}(\hat{\alpha}) = \mathbb{E}_{\mathcal{X}}[\frac{\partial \log p_n(\mathcal{X};\hat{\alpha})}{\partial \hat{\alpha}_i}  \frac{\partial \log p_n(\mathcal{X};\hat{\alpha})}{\partial \hat{\alpha}_j}].
    \label{eq13}
    \end{equation}
    Remembering that $N$ denotes the number of classes as described in Eq.~\ref{eq:nas_objective}. It is then trivial to prove that the Fisher information of the probability distribution set $P$ approaches a scaled version of the Hessian of log-likelihood as
    \begin{equation}
        I_{i,j}(\hat{\alpha}) = -\mathbb{E}_{\mathcal{X}}[ \frac{\partial^2 \log p_n(\mathcal{X};\hat{\alpha})}{\partial \hat{\alpha}_i \partial \hat{\alpha}_j} ].
    \label{eq14}
    \end{equation}
    
    Let $H_{i,j}$ denotes the second-order partial derivatives $\frac{\partial^2}{\partial \hat{\alpha}_i \partial \hat{\alpha}_j}$. Noting that the first derivatives of log likelihood is 
    \begin{equation}
        \frac{\partial \log p_n(\mathcal{X};\hat{\alpha})}{\partial \hat{\alpha}_i}  = \frac{\partial p_n(\mathcal{X};\hat{\alpha})}{p_n(\mathcal{X};\hat{\alpha}) \partial \hat{\alpha}_i},
    \label{eq15}
    \end{equation}
    the second derivatives is 
    \begin{equation}
        {\bf H}_{i,j} \log p_n(\mathcal{X};\hat{\alpha}) = \frac{{\bf H}_{i,j} p_n(\mathcal{X};\hat{\alpha})}{p_n(\mathcal{X};\hat{\alpha})} - \frac{\partial p_n(\mathcal{X};\hat{\alpha})}{p_n(\mathcal{X};\hat{\alpha}) \partial \hat{\alpha}_i} \frac{\partial p_n(\mathcal{X};\hat{\alpha})}{p_n(\mathcal{X};\hat{\alpha}) \partial \hat{\alpha}_j}.
    \label{eq16}
    \end{equation}
    Considering that 
    \begin{equation}
    \begin{aligned}
        \mathbb{E}_{\mathcal{X}}(\frac{{\bf H}_{i,j}p_n(\mathcal{X};\hat{\alpha})}{p_n(\mathcal{X};\hat{\alpha})}) &= \int \frac{{\bf H}_{i,j}p_n(\mathcal{X};\hat{\alpha})}{p_n(\mathcal{X};\hat{\alpha})}p_n(\mathcal{X};\hat{\alpha})d\mathcal{X} \\
        &= {\bf H}_{i,j}\int p_n(\mathcal{X};\hat{\alpha})d\mathcal{X} = 0, \\
    \end{aligned}
    \label{eq17}
    \end{equation}
    we take the expectation of the second derivative and then obtain
    \begin{equation}
    \begin{aligned}
        &\mathbb{E}_{\mathcal{X}}({\bf H}_{i,j}\log p_n(\mathcal{X};\hat{\alpha})) = - \mathbb{E}_{\mathcal{X}}\{ \frac{\partial p_n(\mathcal{X};\hat{\alpha})}{p_n(\mathcal{X};\hat{\alpha}) \partial \hat{\alpha}_i} \frac{\partial p_n(\mathcal{X};\hat{\alpha})}{p_n(\mathcal{X};\hat{\alpha}) \partial \hat{\alpha}_j}\} \\ 
        &= -\mathbb{E}_{\mathcal{X}}\{ \frac{\partial p_n(\mathcal{X};\hat{\alpha})}{\partial \hat{\alpha}_i} \frac{\partial p_n(\mathcal{X};\hat{\alpha})}{\partial \hat{\alpha}_j}\}.
    \end{aligned}
    \label{eq18}
    \end{equation}
    Thus an equivalent substitution for the Hessian matrix ${\bf H}_{\tilde{f_{\bf b}}}(\hat{\alpha})$ in Eq.~\ref{eq9} is the product of two first-order derivatives. This concludes the proof that we can use the covariance of gradients to represent the Hessian matrix for efficient computation. 
    
    \subsection{Decoupled optimization for training the DCP-NAS}
    \re{As one of the critical steps to achieve a powerful NAS strategy, the optimization of $\alpha$ plays an essential role. However, existing NAS and BinaryNAS methods neglect the intrinsic coupling relationship of the weights and architecture parameters, resulting in a sub-optimal architecture caused by an insufficient searching process.
    }
    In this section, we first describe the coupling relationship between the weights and architecture parameters existing in the DCP-NAS. Then we present the decoupled optimization during backpropagation of the sampled supernet for fully and effectively optimizing such two coupling parameters.  
    
    \subsubsection{Coupled models for DCP-NAS}
\begin{figure}[t]
		\centering
		\includegraphics[scale=.55]{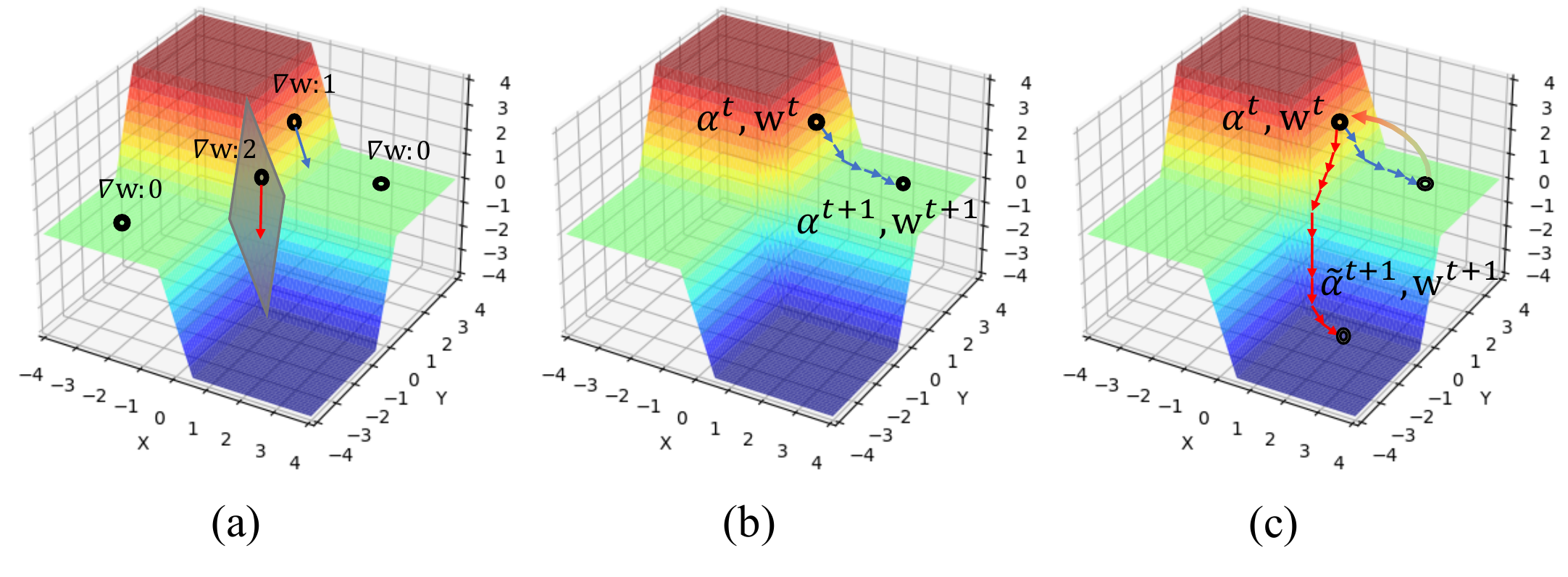}
		\caption{The loss landscape illustration of supernet. (a) The gradient of current weights with different $\alpha$, (b) The vanilla $\alpha^{t+1}$ with backpropagation, (c) $\tilde{\alpha}^{t+1}$ with the decoupled optimization.}
		\label{fig:decouple}
	\end{figure}

    Combing Eq.~\ref{eq:node} and Eq.~\ref{eq:soft_performance}, we first show how parameters in DCP-NAS are formulated in a coupling relationship as ${\bf a}^{(j)} = \sum_{i<j} \operatorname{softmax}(\alpha_m^{(i,j)}) ({\bf w}^{(i,j)} \otimes {\bf a}^{(i)})$, where ${\bf w}^{(i,j)} = [[{\bf w}_m]] \in \mathbb{R}_{M \times 1}, {\bf w}_m \in \mathbb{R}_{C_{out}\times C_{in}\times K_m \times K_m}$ denote the weights of all candidate operations between $i$-th and $j$-th nodes and $K_m$ denotes the kernel size of $m$-th operation. Specifically, for pooling and identity operations, $K_m$ equals to the downsample size and the feature map size, ${\bf w}_m$ equals to ${\bf 1} / (K_m \times K_m)$ and ${\bf 1}$, respectively. 
    For each intermediate node, its output ${\bf a}^{(j)}$ are jointly determined by $\alpha_m^{(i,j)}$ and ${\bf w}_m^{(i,j)}$, while ${\bf a}^{(i)}$ is independent with both $\alpha_m^{(i,j)}$ and ${\bf w}_m^{(i,j)}$. {As shown in Fig~\ref{fig:decouple} (a) and (b), with different $\alpha$, the gradient of corresponding ${\bf w}$ can be various and sometimes hard to be optimized, which are possibly trapped into a local minima. However, with decouple the $\alpha$ and ${\bf w}$, the supernet has the chance to jump out of a local minima and be optimized with a better convergence. }
    
    Based on the deviation and analysis above, we propose our objective for optimizing the neural architecture search process
    \begin{equation}\label{eq:coupled_objective}
    \begin{aligned}
        \arg\min_{\alpha, {\bf w}} \mathcal{L}({\bf w}, \alpha) = 
        \left\{
        \begin{aligned}
        &\mathcal{L}_{\operatorname{NAS}} + {reg}({\bf w}), &\operatorname{for \;Parent \;model}\\
        &\mathcal{L}_{\operatorname{DCP-NAS}} + {reg}({\bf w}), &\operatorname{for \;Child \;model\;\;}     
        \end{aligned}
        \right.
    \end{aligned}
    \end{equation}
    where $\alpha \in \mathbb{R}_{E \times M}, {\bf w} \in \mathbb{R}_{M \times 1}$ and ${reg}({\cdot})$ denotes the regularization item. Following~\cite{liu2018darts,xu2019pcdarts}, the weights ${\bf w}$ and the architectural parameters $\alpha$ are optimized sequentially, in which ${\bf w}$ and $\alpha$ are updated independently. However, it is improper to optimize ${\bf w}$ and $\alpha$ independently due to their coupling relationship. 
    We consider the searching and training process of differentiable Chile-Parent neural architecture search as a coupling optimization problem and solve the problem using a new backtracking method. The details will be shown in Section~\ref{sec:decouple}.

    \subsubsection{Decoupled optimization for Child-Parent model}\label{sec:decouple}
    We reconsider the coupling relation between ${\bf w}$ and $\alpha$ from a new perspective. The derivative calculation process of ${\bf w}$ should consider its coupling parameters $\alpha$. Based on the chain rule~\cite{petersen2008matrix} and its notations, we have
    \begin{equation}
        \re{\begin{aligned}
            \tilde{\alpha}^{t+1} &= \alpha^t + \eta_1(-\frac{\partial \mathcal{L}(\alpha^t, {\bf w}^t)}{\partial \alpha^t} + \eta_2 Tr[(\frac{\partial \mathcal{L}(\alpha^t, {\bf w}^t)}{\partial {\bf w}^t})^T \frac{\partial {\bf w}^t}{\partial \alpha^t}]) \\
            &= \alpha^{t+1} + \eta_1 \eta_2 Tr[(\frac{\partial \mathcal{L}(\alpha^t, {\bf w}^t)}{\partial {\bf w}^t})^T \frac{\partial {\bf w}^t}{\partial \alpha^t}],
        \end{aligned}}
    \label{decouple}
    \end{equation}
    where $\eta_1$ represents the learning rate, $\eta_2$ represents the coefficient of backtracking, $\tilde{\alpha}^{t+1}$ denotes the value after backtracking from the vanilla $\alpha^{t+1}$, while vanilla $\alpha^{t+1}$ is calculated from the backpropagation rule and corresponding optimizer in the neural network. $Tr(\cdot)$ represents the trace of a matrix. 
    However, the item $\frac{\partial {\bf w}^t}{\partial \alpha^t}$ of Eq. \ref{decouple} is undefined and unsolvable based on the normal backpropagation process.
    To address this problem, we propose a decoupled optimization method as following. In the following, we omit the superscript $\cdot^t$ and define $\tilde{\mathcal{L}}$ as
    \begin{equation}\label{eq:hat_G}
        \tilde{\mathcal{L}} = (\frac{\partial \mathcal{L}(\alpha, {\bf w})}{\partial {\bf w}})^T / \alpha, 
    \end{equation}
    which considers the coupling optimization problem as in Eq.~\ref{eq:coupled_objective}. Note that $R(\cdot)$ is only considered when backtracking. Thus we have
    \begin{equation}\label{eq:partial}
        \frac{\partial \mathcal{L}(\alpha, {\bf w})}{\partial {\bf w}} = Tr[\alpha\tilde{\mathcal{L}}\frac{\partial {\bf w}}{\partial \alpha}]. 
    \end{equation}
    For simplifying the derivation, we rewrite $\tilde{\mathcal{L}}$ as $[\tilde{g}_1, \tilde{g}_e, \cdots, \tilde{g}_E]$, where each $\tilde{g}_e$ is a column vector. Assuming that ${\bf w}_m$ and $\alpha_{i,j}$ are independent when $m \neq j$, $\alpha_{i,j}$ denotes an specific element in matrix $\alpha$, we have
    \begin{equation}\label{eq:matrix_1}
		(\frac{\partial {\bf w}}{\partial \alpha})_m =
		\begin{bmatrix} 
		0 &...&\frac{\partial {\bf w}_m}{\partial \alpha_{1, m}}&...& 0 \\
		.&&.&&.\\
		0 &...&\frac{\partial {\bf w}_m}{\partial \alpha_{e, m}}&...& 0 \\
		.&&.&&.\\ 
		0 &...&\frac{\partial {\bf w}_m}{\partial \alpha_{E, m}}&...& 0
		\end{bmatrix}_{E \times M}
	\end{equation} 
    and with rewritten $\alpha$ as a column vector $[\alpha_1, \alpha_e , \cdots, \alpha_E]^T$ with each $\alpha_e$ is a row vector, we have
	\begin{equation}\label{eq:matrix_2}
		\alpha \tilde{\mathcal{L}} =
		\begin{bmatrix} 
		\alpha_1 \tilde{g}_1 &...&\alpha_1 \tilde{g}_e&...& \alpha_1 \tilde{g}_E \\
		.&&.&&.\\
		\alpha_e \tilde{g}_1 &...&\alpha_e \tilde{g}_e&...& \alpha_e \tilde{g}_E \\
		.&&.&&.\\ 
		\alpha_E \tilde{g}_1 &...&\alpha_E \tilde{g}_e&...& \alpha_E \tilde{g}_E \\
		\end{bmatrix}_{E \times E}. 
	\end{equation} 
    Combing Eq.~\ref{eq:matrix_1} and Eq.~\ref{eq:matrix_2}, the matrix in the trace item of Eq.~\ref{eq:hat_G} can be written as
	\begin{equation}\label{eq:matrix_hat_G}
	    \alpha\tilde{\mathcal{L}}(\frac{\partial {\bf w}}{\partial \alpha})_m = 
	    \begin{bmatrix} 
		0 &...&\alpha_1 \sum_{e'=1}^E \tilde{g}_{e'} \frac{\partial {\bf w}_m}{\partial \alpha_{e', m}} &...& 0 \\
		.&&.&&.\\
		0 &...&\alpha_e \sum_{e'=1}^E \tilde{g}_{e'} \frac{\partial {\bf w}_m}{\partial \alpha_{e', m}} &...& 0 \\
		.&&.&&.\\ 
		0 &...&\alpha_E \sum_{e'=1}^E \tilde{g}_{e'} \frac{\partial {\bf w}_m}{\partial \alpha_{e', m}} &...& 0 \\
		\end{bmatrix}_{E \times M}. 
	\end{equation}
    Thus the whole matrix $\alpha\tilde{\mathcal{L}}(\frac{\partial {\bf w}}{\partial \alpha})$ is with the size of $E \times M \times M$. After the above derivation, we compute the $e$-th component of the trace item in Eq.~\ref{eq:hat_G} as
	\begin{equation}\label{eq:trace}
	    Tr[\alpha\tilde{\mathcal{L}}(\frac{\partial {\bf w}}{\partial \alpha})]_e = \alpha_e \sum_{m=1}^M \sum_{e'=1}^E \tilde{g}_{e'} \frac{{\bf w}_m}{\partial \alpha_{e', m}} 
	\end{equation}
    Noting that in the vanilla propagation process, $\alpha^{t+1} = \alpha^t - \eta_1\frac{\partial \mathcal{L}(\alpha^t)}{\partial \alpha^t}$, thus combining Eq.~\ref{eq:trace} we have
	\begin{equation}\label{eq:hat_alpha}
		\begin{aligned}
		\tilde{\alpha}^{t+1} &= \alpha^{t+1} - \eta  
		\begin{bmatrix} 
		\sum_{m=1}^M \sum_{e'=1}^E \tilde{g}_{e'} \frac{\partial {\bf w}_m}{\partial \alpha_{e', m}}\\
		.\\
		\sum_{m=1}^M \sum_{e'=1}^E \tilde{g}_{e'} \frac{\partial {\bf w}_m}{\partial \alpha_{e', m}}\\
		.\\ 
		\sum_{m=1}^M \sum_{e'=1}^E \tilde{g}_{e'} \frac{\partial {\bf w}_m}{\partial \alpha_{e', m}}
		\end{bmatrix}
		\circledast
		\begin{bmatrix} 
		\alpha_1\\
		.\\
		\alpha_e\\
		.\\ 
		\alpha_E
		\end{bmatrix}\\    
		&= \alpha^{t+1} + \eta \psi^t \circledast \alpha^t,	
		\end{aligned}
	\end{equation}
    where $\circledast$ represents the Hadamard product and $\eta = \eta_1 \eta_2$. We take $\psi^t = -[\sum_{m=1}^M \sum_{e'=1}^E \tilde{g}_{e'} \frac{\partial {\bf w}_m}{\partial \alpha_{e', m}}, \cdots, \sum_{m=1}^M \sum_{e'=1}^E \tilde{g}_{e'} \frac{\partial {\bf w}_m}{\partial \alpha_{e', m}}]^T$. Note that, $\frac{\partial {\bf w}}{\partial \alpha}$ is unsolvable and has no explicit form in NAS, which causes a unsolvable $\psi^t$. Thus we introduce a learnable parameter $\tilde{\psi}^t$ for approximating $\psi^t$, which back-propagation process is calculated as
    \begin{equation}\label{eq:gamma}
        \tilde{\psi}^{t+1} = | \tilde{\psi}^t - \eta_{\psi} \frac{\partial \mathcal{L}}{\partial \tilde{\psi}^t} |. 
    \end{equation}
	
    Eq.~\ref{eq:hat_alpha} shows our method is actually based on a projection function to solve the coupling problem of the optimization by the learnable parameter $\tilde{\psi}^t$. In this method, we consider the influence of $\alpha^t$ and backtrack the optimized state at the $(t + 1)$-th step to form $\tilde{\alpha}^{t+1}$.  However, where and when the backtracking should be applied is the key point in the optimization, thus we define the update rule as
	\begin{equation}\label{eq:update_rule}
	    \tilde{\alpha}_m^{t+1} = \left\{ \begin{matrix}
	    P(\alpha_{:, m}^{t+1}, \alpha_{:, m}^t), & & \operatorname{if} ranking(R({\bf w}_m)) > \tau \\
	    \tilde{\alpha}_{:, m}^{t+1}, & & \operatorname{otherwise}  \\
	    \end{matrix} \right.
	\end{equation}
    where $P(\alpha_{:, m}^{t+1}, \alpha_{:, m}^t) = \alpha_{:, m}^{t+1} + \eta \tilde{\psi}^t \circledast \alpha_{:, m}^t$ and subscript $\cdot_m$ denotes a specific edge. $R({\bf w}_m)$ denotes the norm constraint of ${\bf w}_m$ and is further defined as 
	\begin{equation}
	    R({\bf w}_m) = \|{\bf w}_m\|_2^2, \;\; \forall{m=1,\cdots,M},
	    \label{last}
	\end{equation}
    where $\tau$ denotes the threshold of deciding or not to backtrack. We further define the threshold as
	
	\begin{equation}\label{eq:threshold}
	    \tau = \lfloor \epsilon \cdot M \rfloor
	\end{equation}
    where $\epsilon$ denotes a hyper-parameter for controlling the percentage of edges backtracking. With backtracking $\alpha$, the supernet can learn to jump out of the local minima. The overall process of DCP-NAS is outlined in Algorithm \ref{algorithm_dcp}. Note that the decoupled optimization can be employed to both Parent and Child model. When applied to Child model, the ${\bf w}$ here denotes the reconstructed weights from the binary weights, {\em i.e.}, ${\bf w} = \beta \circ {\bf b}^{\hat {\bf w}}$. 
	
	\begin{algorithm}[t]
	    \caption{Search process of DCP-NAS}
	    {\bf Input:} Training data, validation data \\
	    {\bf Parameter:} Searching hyper-graph: $\mathcal{G}, M = 8, e(o_m^{(i,j)}) = 0$ for all edges \\
	    {\bf Output:} Optimized $\hat{\alpha}^*$.
	    \begin{algorithmic}[1]
	    \WHILE{DCP-NAS}
		    \WHILE{Training real-valued Parent}
		    \STATE Search a temporary real-valued architecture $p({\bf w},\alpha)$.
		    \STATE Decoupled optimization from Eqs.~\ref{decouple} to~\ref{last}.
		    \STATE Generate the tangent direction $\frac{\partial \tilde{f}({\bf w}, \alpha)}{\partial \alpha}$ from Eqs.~\ref{eq:nas_objective} to~\ref{eq4}.
		    \ENDWHILE
		    \STATE Architecture inheriting $\hat{\alpha}\leftarrow\alpha$.
		    \WHILE{Training 1-bit Child}
		    \STATE Calculate the learning objective from Eqs.~\ref{eq5} to~\ref{eq9}.
		    \STATE Tangent propagation from Eqs.~\ref{eq10} to~\ref{eq18} and decoupled optimization from Eqs.\\~\ref{decouple} to~\ref{last}.
		    \STATE Obtain the $\hat{p}(\hat{{\bf w}},\hat{\alpha})$.
		    \ENDWHILE
		    \STATE Architecture inheriting ${\alpha}\leftarrow\hat{\alpha}$.
	    \ENDWHILE
	    \STATE {\bf return} Optimized architecture $\hat{\alpha}^*$.
	    \end{algorithmic}
    \label{algorithm_dcp}
    \end{algorithm}

    \section{Experiments}
    We quantitatively demonstrate the effectiveness of our DCP-NAS in this section. We first provide detailed ablation studies of the proposed DCP-NAS in Section~\ref{ablation}. Then we evaluate the proposed DCP-NAS on the architecture search for image classification on the widely-used CIFAR-10~\cite{krizhevsky2014cifar} and ImageNet ILSVRC12~\cite{deng2009imagenet} datasets with different search spaces and constraints in Section~\ref{sec:classification}. Finally, to further validate the effectiveness and generalizability, we transfer the architectures searched on ImageNet to person re-identification and object detection task from Sections~\ref{sec:identification} to \ref{sec:detection}. 
    In the following experiments, both the kernels and the activations are binarized. The leading performances reported in the following sections verify the superiority of our DCP-NAS.

    \subsection{Datasets and Implementation Details}\label{datasets}
    \subsubsection{Datasets}
    CIFAR-10~\cite{krizhevsky2014cifar} is a natural image classification dataset, which is composed of a training set and a test set,  with 50,000 and 10,000 32$\times$32 color images, respectively. These images span 10 different classes, including airplanes, automobiles, birds, cats, deer, dogs, frogs, horses, ships, and trucks. 
    ImageNet ILSVRC12 object classification dataset~\cite{deng2009imagenet} is more diverse and challenging. It contains 1.2 million training images, and 50,000 validation images, across 1000 classes.
	
    The PASCAL VOC dataset contains natural images from 20 different classes. We train our model on the VOC {\tt trainval2007} and VOC {\tt trainval2012} sets, which consist of approximately 16k images. We then evaluate our method on the VOC {\tt test2007} set, which includes 4952 images. Following~\cite{everingham2010pascal}, we use the mean average precision (mAP) as the evaluation criterion.

    The COCO dataset consists of images from 80 categories. We conduct experiments on the COCO {\tt 2014}~\cite{lin2014microsoft} object detection track. Models are trained with the combination of 80k images from the COCO {\tt train2014} and 35k images sampled from COCO {\tt val2014}, {\em i.e.}, COCO {\tt trainval35k}. On the remaining 5k images from the COCO {\tt minival}, we test our method based on the average precision (AP) for IoU$\in$ [0.5 : 0.05: 0.95] denoted as mAP@[.5, .95]. We also report AP$_{50}$, AP$_{75}$, AP$_s$, AP$_m$, and AP$_l$ to further analyze our method.
    
    The Market-1501~\cite{zheng2015scalable}, DukeMTMC-reID~\cite{zheng2017discriminatively} and CUHK03~\cite{li2014deepreid} are amoong the largest Re-ID benchmark datasets based on images. In Market-1501, 32,668 labeled bounding boxes of 1,501 identities captured from six different perspectives are contained, which are detected using Deformable Part Model (DPM)~\cite{felzenszwalb2009object}. The Market-1501 dataset is split into two parts: a training dataset consisting of 12,936 images with 751 identities and a test dataset containing 19,732 images with 750 identities. For testing, 3,368 hand-painted images with 750 identities are used as probe set to classify the identities of test dataset. DukeMTMC-reID is a subset of the DukeMTMC for Re-ID based on images, with the same format as the Market-1501 dataset. The original dataset is sampled from 85-minute high-resolution videos recorded by eight different cameras. Hand-painted pedestrian bounding boxes are also available for classification during test. CUHK03 offers both hand-labeled and DPM-detected bounding boxes, and we use the latter in this article. CUHK03 originally adopts 20 random train/test splits, which is time-consuming for deep learning.
	
    \subsubsection{Search protocol}
	
    In the search process, we consider a total of 6 cells with the initial 16 channels in the network, where the reduction cell are inserted in the second and the fourth layers, and the others are normal cells. There are 4 intermediate nodes in each cell. The initial number of operations $M$ is set as 8 and the number of search epochs is set as 105 (same as CP-NAS~\cite{zhuo2020cp}) for fairness. $\lambda$ is set as $1e-4$, and the batch size is set to 512. We use Adam with momentum to optimize the network weights, with an initial learning rate of 0.025 (annealed down to zero following a cosine schedule), a momentum of 0.9, and a weight decay of $5e-4$. When we search for the architecture directly on ImageNet, we use the same parameters for searching with CIFAR-10 except that the initial learning rate is set to 0.05 and $\lambda$ is set to $1e-3$. Due to the efficient guidance of DCP-NAS model, we use $50\%$ of the training set with CIFAR-10 and ImageNet for architecture search and $5\%$ of the training set for evaluation, leading to a faster search. After search, in the architecture evaluation step, our experimental settings are similar to~\cite{liu2018darts,zoph2018learning,pham2018efficient}.
	
    \subsubsection{Training protocol}
    A larger network of 10 cells (8 normal cells and 2 reduction cells) is trained on CIFAR-10 for 600 epochs with a batch size of 96 and an additional regularization cutout~\cite{devries2017improved}. The initial number of channels is set as 54, 72, 108 for different model sizes, {\em i.e.}, DCP-NAS-S/M/L in this paper. We use the Adam optimizer with an initial learning rate of 0.0025 (annealed down to zero following a cosine schedule without restart), a momentum of 0.9, a weight decay of $1e-5$, and a gradient clipping at 5. 

    When stacking the cells to evaluate on ImageNet, the evaluation stage follows that of DARTS~\cite{liu2018darts}, which starts with three convolutional layers with s stride of 2 to reduce the input image resolution from $224 \times 224$ to $28 \times 28$. 10 cells (8 normal cells and 2 reduction cells) are stacked after these three layers, with the initial channel number being 54, 72, 108 for different model sizes, {\em i.e.}, DCP-NAS-S/M/L. The network is trained from scratch for 512 epochs using a batch size of 512. We use Adam optimizer with a momentum for 0.9, an initial learning rate of 0.001 (decayed down to zero following a cosine schedule), and a weight decay of $3e-5$. Further enhancements are adopted including label smoothing and an auxiliary loss tower during training. All the experiments and models are implemented in PyTorch~\cite{paszke2017automatic}.
    
    On person re-identification task, we utilize SGD as the optimizer with momentum set as 0.9 and weight decay set as $1e-4$. On all three datasets, initial learning rates are both set to 0.05 with a Cosine Annealing learning rate decay. The total epochs are 120 and the batch size is set as 64. For object detection task, we implement experiments following~\cite{xu2021layer} on two-stage Faster-RCNN with FPN neck, the lateral connection of the FPN~\cite{lin2017feature} neck is replaced with $3 \times 3$ 1-bit convolution for improving performance. This adjustment is implemented in all of the Faster-RCNN experiments. For Faster-RCNN, we train the model for 12 epochs at a learning rate of 0.02, which decays by multiplying 0.1 in the 9-th and 11-th epochs.
 	
    \begin{table}[tb]
    \caption{Effect of with/without the reconstruction error and tangent direction constraint on the ImageNet dataset. The architecture conducted the experiments is DCP-NAS-L.}
    \centering
    \begin{tabular}{c c c c c c }
    \toprule
    \multicolumn{2}{c}{Tangent direction (${\bf D}(\hat{\alpha})$)} & \XSolidBrush  & \CheckmarkBold & \XSolidBrush  & \CheckmarkBold   \\ \cmidrule(lr){3-6}
    \multicolumn{2}{c}{Reconstruction error ($\mathcal{L}_R(\hat{\bf w}, \beta)$)} & \XSolidBrush & \XSolidBrush & \CheckmarkBold & \CheckmarkBold  \\ \hline
    \multirow{2}{*}{Accuracy}& Top-1 & 66.7 & 68.3 & 68.2 & \textbf{72.4} \\ \cmidrule(lr){2-6}
    & Top-5 & 83.3 & 85.0 & 85.1 &\textbf{89.2} \\ \bottomrule
    \end{tabular}
    \label{table:tangent}
    \end{table}

	\begin{figure*}[tb]
		\centering
		\includegraphics[width=0.7\linewidth]{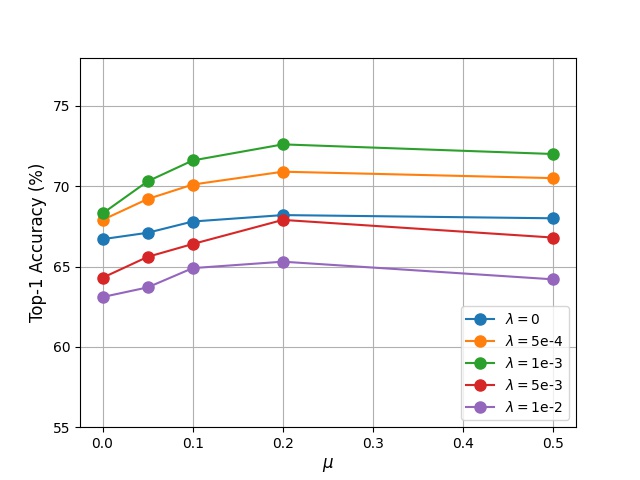}
		\caption{With different $\lambda$ and $\mu$, we evaluate the Top-1 accuracies of DCP-NAS-L} on ImageNet. 
		\label{figure:lambda}
	\end{figure*}
	
	\subsection{Ablation Study}
	\label{ablation}
	
    \subsubsection{Effectiveness of Tangent Propagation}
    In this section, we evaluate the effects of the tangent propagation on the performance of DCP-NAS, the hyper-parameter used in this section includes $\lambda$, $\mu$. Furthermore, we also discuss the effectiveness of the reconstruction error. The implementation details are given below. 
	
    For searching the binary neural architecture in a better form, $\lambda$ and $\mu$ are used to balance the KL divergence $\tilde{f}(\hat{\bf w}, \hat{\alpha}, \beta)$ for supervising the Child, the reconstruction error for binary weights $\mathcal{L}_R(\hat{\bf w}, \beta)$ and the tangent direction constraint ${\bf D}(\hat{\alpha})$. We evaluate $\lambda$ and $\mu$ on ImageNet dataset with the architecture DCP-NAS-L. To better understand the tangent propagation on the large-scale ImageNet ILSVRC12 dataset, we experiment to examine how the tangent direction constraint affects performance. According to the above experiments, we first set $\lambda$ to $5e-3$ and $\mu$ to $0.2$, if they are used. As shown in Tab.~\ref{table:tangent}, both the tangent direction constraint and the reconstruction error can independently improve the accuracy on ImageNet. When applied together, the Top-1 accuracy reaches the highest value of 72.4\%. Then we conduct experiments with various value of $\lambda$ and $\mu$ as shown in Tab.~\ref{figure:lambda}. We observe that with a fixed value of $\mu$, the Top-1 accuracy increases at the beginning with $\lambda$ increasing, but decreases when $\lambda$ is greater than 1e-3. For when $\lambda$ becomes larger, DCP-NAS tends to select the binary architecture with similar gradient to its real-valued counterpart. To some extent, the accuracy of \re{1-bit} model is neglected , which leads to a performance drop. Another performance variation phenomenon is that the Top-1 accuracy rise first and then fall with the increase of $\mu$ while $\lambda$ containing fixed values. For too muach attention paid to minimizing the distance between \re{1-bit} parameters and their counterparts may introduces a representation ability collapse to the \re{1-bit} models and severely degenerates the performance of DCP-NAS.

    \begin{table}[]
    \centering
    \caption{We compare the searching efficiency of different search strategies on ImageNet, including previous NAS on both real-valued and \re{1-bit} search space, random search and our DCP-NAS. T.P. and D.O. denote Tangent Propagation and Decoupled Optimization, respectively. }
    \begin{tabular}{cc|c|c|c|c|c}
    \hline
    \multicolumn{2}{c|}{Method}                                                                                                                                   & T.P.                          & GGN                           & D.O.                          & \begin{tabular}[c]{@{}c@{}}Top-1 Acc\\ (\%)\end{tabular} & \begin{tabular}[c]{@{}c@{}}Search Cost\\ (GPU days)\end{tabular} \\ \hline
    \multicolumn{1}{c|}{\multirow{3}{*}{\begin{tabular}[c]{@{}c@{}}Real-valued \\ NAS\end{tabular}}} & PNAS                                                        & -                             & -                             & -                             & 74.2                                                     & 225                                                              \\
    \multicolumn{1}{c|}{}                                                                           & DARTS                                                       & -                             & -                             & -                             & 73.1                                                     & 4                                                                \\
    \multicolumn{1}{c|}{}                                                                           & PC-DARTS                                                    & -                             & -                             & -                             & 75.8                                                     & 3.8                                                              \\ \hline
    \multicolumn{1}{c|}{\multirow{3}{*}{\begin{tabular}[c]{@{}c@{}}Direct \\ BNAS\end{tabular}}}    & BNAS$_1$                                                    & -                             & -                             & -                             & 64.3                                                     & 2.6                                                              \\
    \multicolumn{1}{c|}{}                                                                           & BNAS$_2$-H                                                  & -                             & -                             & -                             & 63.5                                                     & -                                                                \\
    \multicolumn{1}{c|}{}                                                                           & Random Search                                               &                               &                               &                               & 51.3                                                     & 4.4                                                              \\ \hline
    \multicolumn{1}{c|}{\multirow{5}{*}{\begin{tabular}[c]{@{}c@{}}Auxiliary \\ BNAS\end{tabular}}} & \begin{tabular}[c]{@{}c@{}}CP-NAS\\ (Baseline)\end{tabular} & -                             & -                             & -                             & 66.5                                                     & 2.8                                                              \\ \cline{2-7} 
    \multicolumn{1}{c|}{}                                                                           & DCP-NAS-L                                                   & \CheckmarkBold & \XSolidBrush   & \XSolidBrush   & 71.4                                                     & 27.9                                                             \\
    \multicolumn{1}{c|}{}                                                                           & DCP-NAS-L                                                   & \CheckmarkBold & \CheckmarkBold & \XSolidBrush   & 71.2                                                     & 2.9                                                              \\
    \multicolumn{1}{c|}{}                                                                           & DCP-NAS-L                                                   & \CheckmarkBold & \XSolidBrush   & \CheckmarkBold & \textbf{72.6}                                                     & 27.9                                                             \\
    \multicolumn{1}{c|}{}                                                                           & DCP-NAS-L                                                   & \CheckmarkBold & \CheckmarkBold & \CheckmarkBold & 72.4                                                     & \textbf{2.9}                                                              \\ \hline
    \end{tabular}
    \label{tab:PCA}
    \end{table}

    \begin{table}[]
    \centering
    \caption{Comparison results on ImageNet dataset with DCP-NAS of distance calculation method used in constraining the gradient of binary NAS under the tangent direction, {\em i.e.}, Eq.~\ref{eq7}. We use the small model size, {\em i.e.}, DCP-NAS-S, to evaluate the searched architecture.}
    \begin{tabular}{ccccc}
    \toprule
    \multirow{2}{*}{Method} & \multicolumn{2}{c}{Accuracy(\%)} & \multirow{2}{*}{\begin{tabular}[c]{@{}c@{}}Memory \\ (MBits)\end{tabular}} & \multirow{2}{*}{\begin{tabular}[c]{@{}c@{}}Search Cost\\ (GPU days)\end{tabular}} \\ \cline{2-3} 
                        & Top-1            & Top-5 & (MBits) & (GPU days)          \\ \hline
    Cosine similarity       & 62.5            & 83.9            & 4.2            & 2.9             \\
    L1-norm                 & 62.7            & 84.3            & 4.3            & 2.9             \\
    \textbf{F-norm}                  & \textbf{63.0}            & \textbf{84.5}            & 4.2            & 2.9             \\ \bottomrule
    \end{tabular}
    \label{tab:norm_eq13}
    \end{table}

    To better understand the acceleration rate of applying Generalized Gauss-Newton matrix (GGN) in the searching process, we conduct experiments to examine the searching cost with and without GGN. As shown in Tab.~\ref{tab:PCA}, we compare the searching efficiency and the accuracy of the architecture obtained by Random Search (random selection), Real-valued NAS methods, Binarized NAS methods, CP-NAS, DCP-NAS without GGN method, DCP-NAS with GGN applied. In random search, the \re{1-bit} supernet randomly samples and trains an architecture in each epoch, then assign the expectation of all performances to each corresponding edges and operations and returns the architecture with the highest score, which lacks the necessary guidance in the searching process, thus has a poor performance for binary architecture search. Notably, our DCP-NAS without GGN are highly computation consumed for the second order gradient is necessarily computed in the tangent propagation. 
    \re{ Note that directly optimizing two supernets is computationally redundant. However, the introduction of GGN for Hessian matrix significantly accelerates the searching process, which reduces the search cost to nearly 10$\%$ with negligible accuracy vibration.} As shown in Table 4, with GGN utilized, our method reduces the search cost from 29 to 2.9, which is more efficient than DARTS .
    Also, our DCP-NAS achieves much lower performance gap between the real-valued NAS with less search cost in a clear margin. To further clarify the tangent propagation, we conduct ablative experiments for different architecture discrepancy calculation methods. 
    As shown in Tab.~\ref{tab:norm_eq13}, F-norm applied in Eq.~\ref{eq7} achieves best performance, while cosine similarity and L1-norm are not so effective as the F-norm.
	
    \begin{figure*}[tb]
        \hspace{-6mm}
        \subfigure[Different $\eta$ with $\epsilon$ = 0.5]{
        \begin{minipage}[t]{0.55\textwidth}
            \centering
            \includegraphics[width= \linewidth]{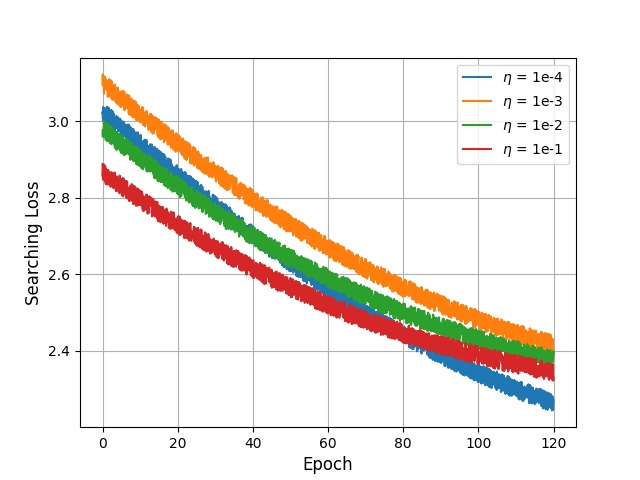}
        \end{minipage}%
        }%
        \hspace{-4mm}
        \subfigure[Different $\epsilon$ with $\eta$ = 1e-4]{
        \begin{minipage}[t]{0.55\textwidth}
            \centering
            \includegraphics[width= \linewidth]{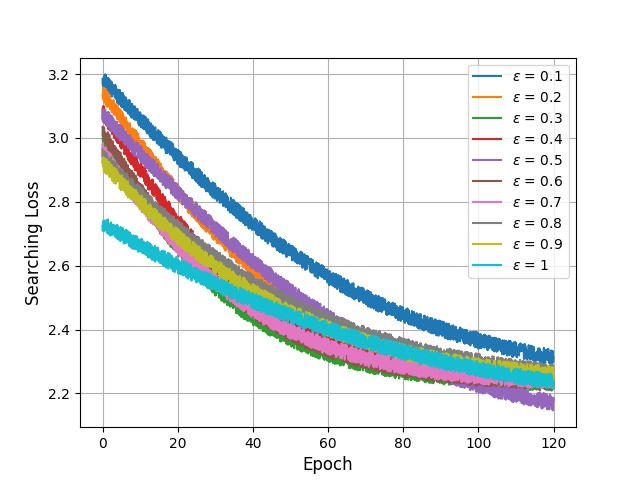}
        \end{minipage}%
        }%
        \caption{Effect of hyper-parameters $\epsilon$ and $\eta$ on one-stage and two-stage training using 1-bit ResNet-18.}
        \label{figure:lr_decoupled}
    \end{figure*}

    \begin{table}[tb]
        \centering
        \caption{Effect of the decoupled optimization on real-valued NAS using ImageNet dataset. D.O. denotes Decoupled Optimization for simplicity.}
        \begin{tabular}{ccccc}
        \toprule
        \multirow{2}{*}{Architecture} & \multicolumn{2}{c}{Accuracy(\%)} & \multirow{2}{*}{\begin{tabular}[c]{@{}c@{}}Memory \\ (MBits)\end{tabular}} & \multirow{2}{*}{\begin{tabular}[c]{@{}c@{}}Search Cost\\ (GPU days)\end{tabular}} \\ \cline{2-3}
                                  & Top-1            & Top-5           &                                                                            &                                                                         \\ \hline
        DARTS                         & 73.1            & 91.0           & 156.8                                                                      & 4                                                                       \\
        \textbf{DARTS + D.O.}                  & \textbf{73.8}            & \textbf{92.1}           & 157.2                                                                      & 4                                                                       \\ \hline
        PC-DARTS                      & 75.8            & 92.7           & 169.6                                                                      & 3.8                                                                     \\ 
        
        \textbf{PC-DARTS + D.O.}               & \textbf{76.6}            & \textbf{93.9}           & 170.3                                                                      & 3.8                                                                     \\ \bottomrule
        \end{tabular}
        \label{tab:do_realnas}
    \end{table}

    \subsubsection{Effectiveness of decoupled optimization}
    In this section, we evaluate the effects of the decoupled optimization for solving the coupling relation between ${\bf w}$ and $\alpha$ in our DCP-NAS. The implementation details are given below. For architecture parameters $\alpha$, we use the Adam optimizer with an initial learning rate $\eta = 1e-4$. We set the the hyper-parameter $\epsilon = 0.2$. The learning rate $\eta_{\psi}$ is set as $1e-4$. As shown in Fig.~\ref{figure:lr_decoupled} (a), when $\eta$ is setted as 1e-4, the searching loss converges to a better minimum. In Fig.~\ref{figure:lr_decoupled} (b), when $\epsilon$ equals to 1, there will be no edges or operations been backtracked through our decoupled optimization, thus the loss in the very beginning of the searching process will be much lower than other conditions. However, the supernet converges to a local minima with higher searching loss at the end of the searching process. Furthermore, when $\epsilon$ is set as 0.1, which denotes that over 90 $\%$ of the edges and operations will be forced to backtrack during optimization. Too much backtracking degenerates the learning of the binary weights and their corresponding scale factors, thus leading to under fitting of the sampled supernet which affects the performance estimation for a better architecture. We also extend our decoupled optimization into real-valued NAS frameworks shown in Tab.~\ref{tab:do_realnas}. For experiments on DARTS~\cite{liu2018darts} and PC-DARTS~\cite{xu2019pcdarts}, our decoupled optimization method further boosts the performance by 0.7\% and 0.8\% with no extra search cost, which proves the effectiveness of our decoupled optimization for real-valued NAS.

    \begin{table*}[tb]
        \centering
        \caption{Performance comparison of architectures obtained under different search spaces. The ``w/o" denotes without in short.}
        \begin{tabular}{ccccc}
        \toprule
        \multirow{2}{*}{Search Space}                                                                       & \multirow{2}{*}{Architecture} & \multicolumn{2}{c}{Accuracy} & Memory  \\
                                                                                                    &                               & Top-1          & Top-5         & (MBits) \\ \hline
        \multirow{3}{*}{Normal search space}                                                                & Small                         & 63.0          & 84.5         & 4.2     \\
                                                                                                    & Medium                        & 69.0          & 88.4         & 6.9     \\
                                                                                                    & Large                         & 72.4          & 90.6         & 6.9     \\ \hline
        \multirow{3}{*}{w/o pooling layer}                                                                  & Small                         & 59.1          & 80.2         & 5.1     \\
                                                                                                    & Medium                        & 64.3          & 85.3         & 7.2     \\
                                                                                                    & Large                         & 68.9          & 87.3         & 7.2     \\ \hline
        \multirow{3}{*}{w/o skip connection}                                                                & Small                         & 60.0          & 81.3         & 5.0     \\
                                                                                                    & Medium                        & 65.1          & 85.9         & 7.1     \\
                                                                                                    & Large                         & 69.5          & 88.6         & 7.1     \\ \hline
        \multirow{3}{*}{\begin{tabular}[c]{@{}c@{}}w/o pooling layer \\ + w/o skip connection\end{tabular}} & Small                         & 58.9          & 80.1         & 5.3     \\
                                                                                                    & Medium                        & 64.0          & 85.1         & 7.4     \\
                                                                                                    & Large                         & 68.2          & 88.9         & 7.4     \\ \bottomrule
        \end{tabular}
        \label{search_space}
    \end{table*}

    \subsubsection{Effectiveness of Search Space}
    In this section, we evaluate the effectiveness of the pooling layer and the skip connection in the search space. 
    We first remove the pooling layer and skip connection from the search space, respectively. Then we simultaneously remove them for further evaluation.
    As shown in the $5$-th $\sim 7$-th rows of Tab.~\ref{search_space}, the performance significantly drops when removing the pooling operations from the search space, primarily due to the insufficient feature fusing.
    Removing all skip-connection operation (identity mapping) from the search space also degrades the network performance, as well as increasing the memory usage to over 1.3$\times$ than the network obtained from the normal search space (described in Sec~\ref{sec:3.3}). 
    As shown in the bottom of Tab.~\ref{search_space}, with removing all pooling and skip-connection operations, the performance of small models drop to only 58.9\% Top-1 accuracy. These experiments show the rationality of our search space setup.

    \subsubsection{Effectiveness of Binarization Proportion}
    
    \begin{table}[]
        \centering
        \caption{Ablative experiments on ImageNet datasets for binarization proportion
        in CP-NAS and DCP-NAS. 
        $^\sharp$ denotes that we ``bianrize activations of preprocessing operations for 2 input nodes". $^\S$ denotes that we ``bianrize activations of the first convolutional layer in the depth-wise separable convolution".}
        \begin{tabular}{ccccc}
        \toprule
        \multirow{2}{*}{Architecture} & Memory  & OPs & \multicolumn{2}{c}{Accuracy (\%)} \\ \cline{4-5} 
                                  & (MBits) & (M) & Top-1            & Top-5            \\ \hline
        DCP-NAS-S$^\sharp$            & 3.5     & 91  & 60.7            & 82.9            \\
        DCP-NAS-S$^\S$                & 3.9     & 98  & 61.2            & 83.2            \\
        \textbf{DCP-NAS-S}                     & 4.2     & 100 & \textbf{63.0}   & \textbf{84.5}   \\ \hline
        DCP-NAS-M$^\sharp$            & 5.8     & 152 & 66.9            & 86.1            \\
        DCP-NAS-M$^\S$                & 6.2     & 159 & 67.2            & 86.7            \\
        \textbf{DCP-NAS-M}                     & 6.9     & 163 & \textbf{69.0}   & \textbf{88.4}   \\ \hline
        CP-NAS-L$^\sharp$             & 11.8    & 307 & 63.1            & 83.0            \\
        CP-NAS-L$^\S$                 & 11.8    & 316 & 64.3            & 84.2            \\
        \textbf{CP-NAS-L}                      & 12.5    & 323 & \textbf{66.5}            & \textbf{86.8}            \\ \hline
        DCP-NAS-L$^\sharp$            & 13.5    & 319 & 69.5            & 88.7            \\
        DCP-NAS-L$^\S$                & 13.8    & 326 & 70.1            & 89.1            \\
        \textbf{DCP-NAS-L}                     & 14.4    & 334 & \textbf{72.4}   & \textbf{90.6}   \\ \bottomrule
        \end{tabular}
        \label{tab:binary_abla}
    \end{table}
    
    \begin{table}[]
        \centering
        \caption{Ablative experiments on the CIFAR-10 and ImageNet datasets for training scheme based on DCP-NAS-S.}
        \begin{tabular}{cccc}
        \toprule
        \multicolumn{2}{c}{Training scheme on CIFAR-10}          & \multicolumn{2}{c}{\multirow{2}{*}{Error (\%)}} \\ \cdashline{1-2}
        Regularization Cutout & Gradient Clipping &                             \\ \hline
        \CheckmarkBold      & \CheckmarkBold  & \multicolumn{2}{c}{\textbf{6.13}}               \\
        \CheckmarkBold      & \XSolidBrush    & \multicolumn{2}{c}{6.42}                        \\
        \XSolidBrush        & \CheckmarkBold  & \multicolumn{2}{c}{6.49}                        \\
        \XSolidBrush        & \XSolidBrush    & \multicolumn{2}{c}{6.87}                        \\ \midrule
        \multicolumn{2}{c}{Training scheme on ImageNet}    & \multicolumn{2}{c}{Accuracy (\%)} \\ \cdashline{1-2}
        Label Smoothing  & Auxiliary Loss   & Top-1                 & Top-5       \\ \hline   
        \CheckmarkBold & \CheckmarkBold & \textbf{63.0}      & \textbf{84.5} \\
        \CheckmarkBold & \XSolidBrush   & 62.9               &  84.3         \\
        \XSolidBrush   & \CheckmarkBold & 62.6               &  84.2         \\
        \XSolidBrush   & \XSolidBrush   & 62.1               &  83.7         \\ \bottomrule
        \end{tabular}
        \label{tab:aug_abla}
    \end{table}
    
    In this section, we evaluate the effects of the binarization of ``activations of the preprocessing operations of the two input nodes" and ``activations of the first convolutional layer in the depth-wise separable convolution". As shown in Tab.~\ref{tab:binary_abla}, with ``activations of the preprocessing operations of the two input nodes" binarized, the Top-1 accuracy decreases up to 2.9\% compare with the final DCP-NAS models, which is less acceptable compared with the marginal memory saving and OPs decrease.
    
    Furthermore, compared with ``activations of the first convolutional layer in the depth-wise separable convolution", the final DCP-NAS achieves desirable performance improvements with acceptable memory saving and OPs. For example, the DCP-NAS-L surpasses DCP-NAS-L$^{\S}$ by 2.3\% Top-1 accuracy, which is significant on ImageNet classification task. Similar phenomenon is also observed on the CP-NAS framework.
    Therefore, our DCP-NAS achieves significant better performance with acceptable extra memory cost, when saving ``activations of the preprocessing operations of the two input nodes" and ``activations of the first convolutional layer in the depth-wise separable convolution" as real-valued. 
    
    \subsubsection{Effectiveness of Training Scheme}
    
    In this section, we evaluate the effects of the training scheme from original CP-NAS~\cite{zhuo2020cp}, {\em i.e.}, additional regularization cutout and gradient clipping used in training DCP-NAS on CIFAR-10, and label smoothing together with auxiliary loss used in training DCP-NAS on ImageNet. As shown in Tab.~\ref{tab:aug_abla}, applying additional regularization cutout and gradient clipping into training on CIFAR-10 brings the best performance, while removing both brings 0.74\% performance drop. 
    Likewise, on ImageNet, applying label smoothing and auxiliary loss into training brings the best performance, which surpasses the counterpart with both removed by 0.9\% Top-1 accuracy.
    Thus, we keep the training protocol consistent with previous works~\cite{zhuo2020cp,liu2018darts,bulat2020bats}. 

    \subsection{Results on Image Classification}\label{sec:classification}
    \begin{table}[tb]
        \caption{Test accuracies on the CIFAR-10 datasets. Quantization method in DCP-NAS are based on MCN~\cite{wang2018modulated}. We calculate the number of parameters for each model, and the numbers refer to the models on CIFAR-10. Err. and Mem. denote Test Error and  Memory Usage, respectively. Gradient-based$^\natural$ denotes our DCP-NAS is a discrepancy-aware gradient-based search method.}
        \centering{
        \begin{tabular}{cccccc}
        \toprule
        \multirow{2}{*}{Architecture} & Err.   & Mem. & W/A  & Search Cost & Search              \\
                                  & (\%)          & (MBits)      &      & (GPU days)  &      Method          \\ \midrule
        WRN-22                        & 5.04          & 138.6         & 32/32 & -           & Manual               \\
        DARTS                         & 2.83          & 108.8          & 32/32 & 4           & Gradient-based       \\
        PC-DARTS                      & 2.78          & 112.0          & 32/32 & 0.15        & Gradient-based       \\ \hline
        WRN-22                        & 5.69          & 4.3         & 1/32 & -           & Manual               \\
        BNAS$_1$                          & 3.94          & 2.6          & 1/32 & 0.09        & Performance-based    \\
        \hline
        CP-NAS-S                & 6.50           & 2.9          & 1/1  & 0.1         & Child-Parent model   \\
        BATS (Small)                & 6.30           & 2.8          & 1/1  & 0.25         & Gradient-based   \\
        \textbf{DCP-NAS-S}       &   \textbf{6.13}            &  2.9            & 1/1  & 0.1         & Gradient-based$^\natural$ \\ \hline
        WRN-22 (BONN)                        & 8.07          & 4.3         & 1/1  & -           & Manual               \\
        \newre{BNAS$_1$ (Large)}                          & 8.29          & 4.5          & 1/1  & 0.09        & \newre{Performance-based}    \\
        \newre{BNAS$_2$ v2-B}  & \newre{6.24}          & \newre{4.5}          & \newre{1/1}  & \newre{0.09}        & \newre{Gradient-based}    \\
        CP-NAS-M               & 5.72          & 4.4          & 1/1  & 0.1         & Child-Parent model   \\
        BATS (Medium)               &     5.60     & 5.4         & 1/1  & 0.25         &  Gradient-based    \\
        \textbf{DCP-NAS-M}      &   \textbf{5.34}            &  4.9            & 1/1  & 0.1         & Gradient-based$^\natural$ \\ \hline
        CP-NAS-L                & 4.73          & 10.6         & 1/1  & 0.1         & Child-Parent model   \\
        BATS (Large)                & 4.50          & 10.0         & 1/1  & 0.25         & Gradient-based   \\
        \textbf{DCP-NAS-L}         & \textbf{4.22} & 11.2         & 1/1  & 0.1         & Gradient-based$^\natural$ \\ \bottomrule
        \end{tabular}
        }
        \label{cifar}
    \end{table}

    \subsubsection{Results on CIFAR-10 datasets}
    We first evaluate our DCP-NAS on CIFAR-10 and compare results with both manually designed networks~\cite{zagoruyko2016wide,he2016deep} and networks searched by NAS at different levels of binarization~\cite{liu2018darts,xu2019pcdarts,chen2020BNAS,zhuo2020cp,bulat2020bats}.

    The results for different architectures on CIFAR-10 are summarized in Tab.~\ref{cifar}. We search for three \re{1-bit CNNs} with different model sizes, which binarize both weights and activations. Note that for the model size, we also consider the number of bits of each parameter. The \re{1-bit CNNs} only need 1 bit to save and compute the weight parameter or the activation, while the real-valued networks need 32.

    Compared with manually designed networks, {\em e.g.}, WRN-22 (BONN)~\cite{zhao2022towards}, our DCP-NAS achieves comparable or smaller test error ($5.34\%$ vs. $8.07\%$) and similar model size ($4.9$MBits vs. $4.3$MBits). Compared with real-valued networks obtained by other NAS methods, our DCP-NAS achieves comparable test errors and significantly more efficient models, with similar or less search time. Compared with state-of-the-art BATS~\cite{bulat2020bats}, our DCP-NAS achieves $0.17\%$/$0.26\%$/$0.28\%$ performance improvements with different model sizes, producing a new state-of-the-art. Moreover, our DCP-NAS consumes $60\%$ lower search cost than BATS. The superior results prove our assumption that our DCP-NAS method can learn a well-performed binary architecture from the tangent space.

    In terms of search efficiency, compared with the previous work~\cite{bulat2020bats}, our DCP-NAS is $60.0\%$ faster (tested on our platform - 8 NVIDIA GEFORCE RTX 2080Ti GPUs). We attribute our superior results to the proposed scheme of GGN-based Hessian matrix acceleration.

    \begin{table}[t]
        \small
        \centering
        \caption{A performance comparison with SOTAs on ImageNet. W/A denotes the bit length of weights and activations. We report the Top-1 (\%)and Top-5 (\%) accuracy. Err. and Mem. denote Test Error and  Memory Usage (MBits), respectively. Search cost here is compared through GPU days. $^\flat$ denotes models trained with an improved training scheme in BNAS$_2$ v2~\cite{kim2021bnas}. $^\ddag$ denotes models trained with an improved training scheme and EBConv in EBN~\cite{bulat2020high}. Gradient-based$^\natural$ denotes our DCP-NAS is a discrepancy-aware gradient-based search method.}
        \hspace{-6mm}
        \setlength{\tabcolsep}{0.8mm}{
        \begin{tabular}{ccccccccc}
        \toprule
        \multirow{2}{*}{Architecture} & \multicolumn{2}{c}{Accuracy(\%)} & \multirow{2}{*}{Mem.} & OPs  & \multirow{2}{*}{W/A} & Search & GPU Mem.  & Search             \\ \cline{2-3}
                                  & Top-1            & Top-5           &                       & (M)  &                      & Cost   & \begin{tabular}[c]{@{}l@{}}\newre{(per-image/}\\ \newre{max Mem.)}\end{tabular}   & Method             \\ \hline
        ResNet-18                     & 69.3            & 89.2           & 357.4                 & 1819 & 32/32                & -      & -      & Manual             \\
        PNAS                          & 74.2            & 91.9           & 163.2                 & 588  & 32/32                & 225    & -      & SMBO               \\
        DARTS                         & 73.1            & 91.0           & 156.8                 & 574  & 32/32                & 4      & \newre{0.01/5.4}    & Gradient-based     \\
        PC-DARTS                      & 75.8            & 92.7           & 169.6                 & 597  & 32/32                & 3.8    & -      & Gradient-based     \\ \hline
        ResNet-18           & \multirow{2}{*}{63.5} & \multirow{2}{*}{85.1} & \multirow{2}{*}{11.2} & \multirow{2}{*}{1819} & \multirow{2}{*}{1/32} & \multirow{2}{*}{-} & \multirow{2}{*}{-}      & \multirow{2}{*}{Manual}             \\
        (PCNN)                        &                 &                &                       &      &                      &        &        &              \\
        BNAS$_1$                      & 71.3            & 90.3           & 6.2                   & 778  & 1/32                 & 2.6    & \newre{0.01/5.4}   & Performance-based  \\ \midrule
        BNAS$_2$-D                    & 57.7            & 79.9           & -                     & 148  & 1/1                  & -      & -      & Gradient-based     \\
        \newre{BNAS$_2$ v2-D}                    & \newre{64.6}            & \newre{84.9}           & \newre{-}                     & \newre{148}  & \newre{1/1}                  & \newre{-}      & \newre{-}      & \newre{Gradient-based}     \\
        BATS                          & 60.4            & 83.0           & -                     & 98   & 1/1                  & -      & -      & Gradient-based     \\
        \textbf{DCP-NAS-S}      & \textbf{63.0}   & \textbf{84.5}  & 4.2                   & 100  & 1/1                  & 2.9    & \newre{0.025/13.1}  & Gradient-based$^\natural$  \\ 
        \textbf{DCP-NAS-S$^\flat$} & \textbf{66.9} & \textbf{87.8} & 4.2 & 100 & 1/1       & 2.9    & \newre{0.025/13.1}  & Gradient-based$^\natural$  \\
        \textbf{DCP-NAS-S$^\ddag$} & \textbf{67.2} & \textbf{88.1} & 4.2 & 100 & 1/1       & 2.9    & \newre{0.025/13.1}  & Gradient-based$^\natural$ \\\midrule
        BNAS$_2$-G                    & 59.8            & 81.6           & -                     & 193  & 1/1                  & -      & -      & Gradient-based  \\ 
        BNAS$_1$                      & 64.3            & 86.1           & 6.4                   & -    & 1/1                  & 3.2    & \newre{0.01/5.4}   & Performance-based  \\ 
        ResNet-18           & \multirow{2}{*}{65.9} & \multirow{2}{*}{86.2} & \multirow{2}{*}{11.2} & \multirow{2}{*}{165} & \multirow{2}{*}{1/1} & \multirow{2}{*}{-} & \multirow{2}{*}{-}      & \multirow{2}{*}{Manual}             \\
        (ReActNet)                    &                 &                &                       &      &                      &        &        &              \\
        BATS          & \multirow{2}{*}{66.1} & \multirow{2}{*}{87.0} & \multirow{2}{*}{-} & \multirow{2}{*}{155} & \multirow{2}{*}{1/1} & \multirow{2}{*}{-} & \multirow{2}{*}{-}      & \multirow{2}{*}{Gradient-based}             \\
        ($2\times$-wider)             &                 &                &                       &      &                      &        &        &              \\
        \re{BARS-D}                      & \re{54.6}       & \re{79.2}      & -                & \re{129}  & \re{1/1}        & \re{-} & \re{-}      & \re{Gradient-based} \\
        \re{BARS-E}                      & \re{56.2}       & \re{80.6}      & -                & \re{183}  & \re{1/1}        & \re{-} & \re{-}      & \re{Gradient-based} \\
        \re{NASB}                      & \re{60.5}       & \re{82.2}      & -                & \re{171}  & \re{1/1}        & \re{-} & \re{-}      & \re{Gradient-based} \\
        \re{NASBV4}                      & \re{65.3}       & \re{85.9}      & -                & \re{281}  & \re{1/1}        & \re{-} & \re{-}      & \re{Gradient-based} \\
        EBN                      & 67.5       & 87.5      & -                & 131  & 1/1        & - & -      & Gradient-based  \\ 
        EBN$^\ddag$              & 71.2       & 90.1      & -                & 131  & 1/1        & - & -      & Gradient-based  \\ 
        \re{EBNAS-L}                      & \re{67.8}       & \re{87.4}      & -                & \re{162}  & \re{1/1}        & \re{-} & \re{-}      & \re{Gradient-based} \\
        \textbf{DCP-NAS-M}     & \textbf{69.0}   & \textbf{88.4}  & 6.9                   & 163  & 1/1                  & 2.9    & \newre{0.025/13.1}  & Gradient-based$^\natural$  \\ 
        \textbf{DCP-NAS-M$^\flat$} & \textbf{72.3} & \textbf{91.1} & 6.9 & 163 & 1/1 & 2.9    & \newre{0.025/13.1}  & Gradient-based$^\natural$  \\
        \textbf{DCP-NAS-M$^\ddag$} & \textbf{72.7} & \textbf{91.3} & 6.9 & 163 & 1/1 & 2.9    & \newre{0.025/13.1}  & Gradient-based$^\natural$  \\\midrule
        BNAS$_2$-H                    & 63.5            & 83.9           & -                     & 656  & 1/1                  & -      & -      & Gradient-based  \\ 
        \re{BARS-F}                      & \re{60.3}       & \re{81.9}      & -                & \re{293}  & \re{1/1}        & \re{-} & \re{-}      & \re{Gradient-based} \\
        \re{NASBV5}                      & \re{66.6}       & \re{87.0}      & -                & \re{352}  & \re{1/1}        & \re{-} & \re{-}      & \re{Gradient-based} \\
        CP-NAS-L                & 66.5            & 86.8           & 12.5                  & 323  & 1/1                  & 2.8    & \newre{0.01/5.4}   & Child-Parent model \\
        \textbf{DCP-NAS-L}      & \textbf{72.4}   & \textbf{90.6}  & 14.4                  & 334  & 1/1                  & 2.9    & \newre{0.025/13.1}  & Gradient-based$^\natural$ \\
        \textbf{DCP-NAS-L$^\flat$} & \textbf{75.0} & \textbf{92.9} & 14.4 & 334  & 1/1     & 2.9    & \newre{0.025/13.1}  & Gradient-based$^\natural$ \\
        \textbf{DCP-NAS-L$^\ddag$} & \textbf{75.2} & \textbf{93.3} & 14.4 & 334  & 1/1     & 2.9    & \newre{0.025/13.1}  & Gradient-based$^\natural$ \\\bottomrule 
        \end{tabular}}
        \label{imagenet}
    \end{table}

    \subsubsection{Results on ImageNet ILSVRC12}
    To further evaluate the performance of our DCP-NAS, we compare our method with state-of-the-art image classification methods on the ImageNet, including hand-crafted ReActNet~\cite{liu2020reactnet}, BATS~\cite{bulat2020bats}, and CP-NAS~\cite{zhuo2020cp}. We also report multi-bit ResNets~\cite{he2016deep} and NAS methods~\cite{cai2018proxylessnas,liu2018darts,xu2019pcdarts,chen2020BNAS} for reference. We use memory usage and operations (OPs)~\cite{liu2018bi} for the efficiency comparison. All the searched networks are obtained directly by DCP-NAS on ImageNet by stacking the cells. All the training settings of $\alpha$ are the same as other weight parameters. The \re{1-bit} models are trained from scratch. Due to the first convolutional layer in a depth-wise separable convolution with fewer parameters, we do not binarize the activations of the first layers for ImageNet. Tab. \ref{imagenet} shows the test accuracy on ImageNet. 
	
    \begin{figure}[t]
        \centering
        \subfigure[The normal cell]{
        \hspace{-26mm}
        \begin{minipage}[t]{1.3\linewidth}
        \centering
        \hspace{19mm}
        \includegraphics[scale=.37]{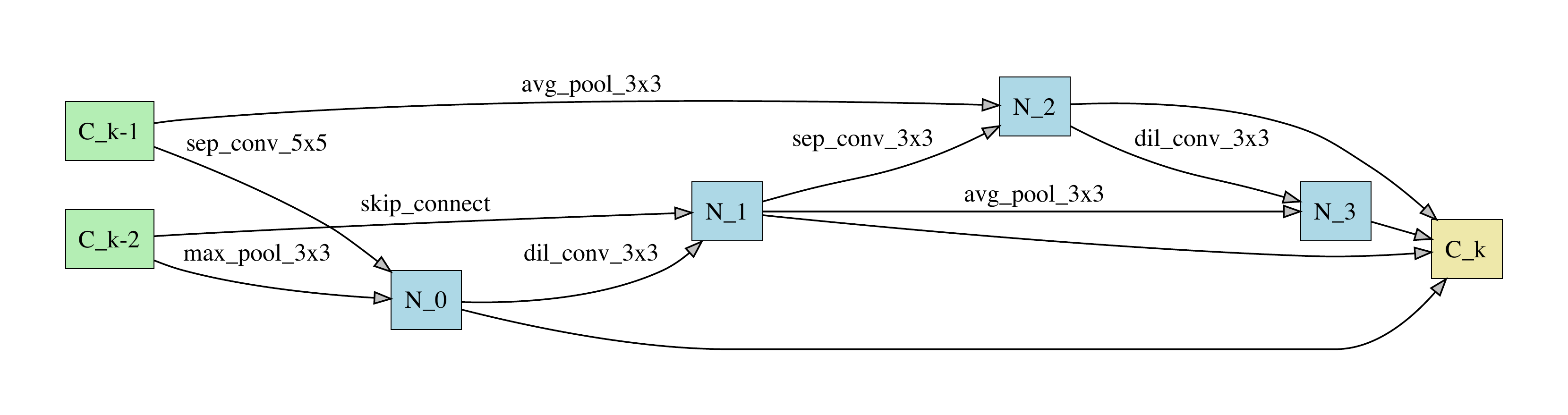}
        \end{minipage}%
        }
        \vspace{-4mm}
        \subfigure[The reduction cell]{
        \begin{minipage}[t]{1.1\linewidth}
        \centering
        \includegraphics[scale=.35]{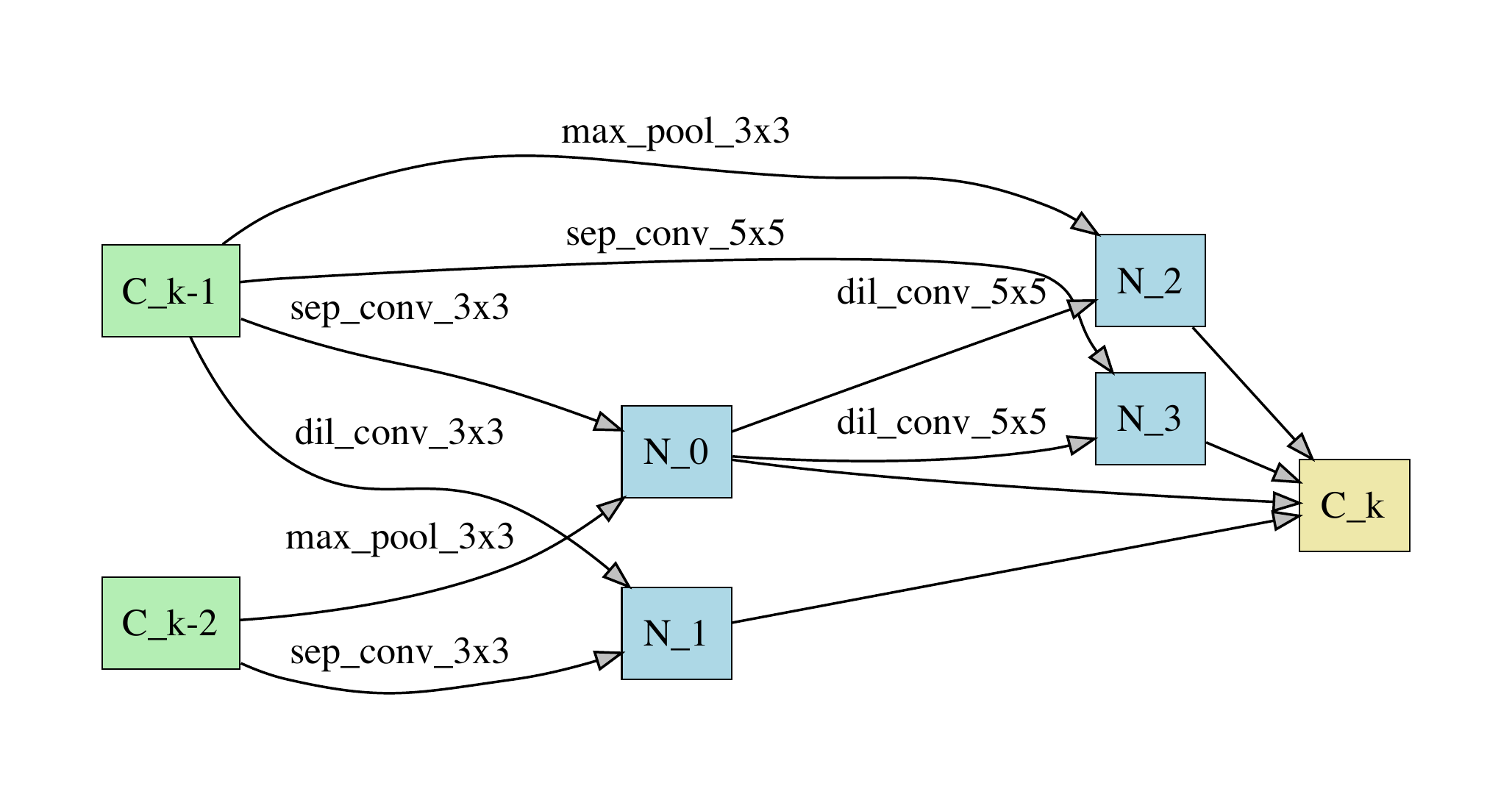}
        \end{minipage}
        }
        \centering
        \caption{The normal cell (a) and the reduction cell (b) searched for ImageNet.}
        \label{cell}
    \end{figure}
	
    \re{We observe that DCP-NAS significantly outperforms previous binary NAS methods with three different model sizes, by significant margins. With the small model, our DCP-NAS improves the Top-1 accuracy by $2.6\%$, $12.8\%$, $3.7\%$, $1.5\%$ and $1.2\%$ compared with BATS, BARS-E, NASBV4, EBN and EBNAS-L. 
    Also, our DCP-NAS improves the Top-1 accuracy by $4.7\%$, and $2.9\%$ compared with BNAS$_1$ and BATS with the medium model. 
    In large networks, DCP-NAS achieves $72.4\%$ vs. $66.5\%$ (CP-NAS). As shown in Fig.~\ref{cell}, our DCP-NAS obtains more optimized architecture for binarized weights and activations, which increase the representation ability of 1-bit CNNs. 
    It is noteworthy that our DCP-NAS-L trained with improved training scheme and EBN following~\cite{bulat2020high} achieves $75.2\%$ top-1 accuracy on ImageNet.}

    Note that compared to the human-designed real-valued networks, our DCP-NAS achieves better performance but with higher compression. Furthermore, to obtain a better performance, we do not binarize the activations of the preprocessing operations for the two input nodes. We achieve an accuracy of $72.4\%$, which surpass the real-valued hand-crafted model, {\em e.g.}, $69.3\%$ for ResNet-18. 

    \begin{table}[t]
        \small
        \renewcommand\arraystretch{1.3}
        \centering
        \caption{Comparison of mAP$(\%)$ and Rank@1$(\%)$ for person re-identification on Market-1501, DukeMTMC-reID, and CUHK03 Datasets. We compare with the BiRe-ID using vanilla ResNets as backbone in~\cite{xu2022bire}. Gradient-based$^\natural$ denotes our DCP-NAS is a discrepancy-aware gradient-based search method.}
        \setlength{\tabcolsep}{1mm}{
        \begin{tabular}{cccccccccc}
        \toprule
        \multirow{3}{*}{Backbone} & \multirow{3}{*}{Quantization} & \multirow{2}{*}{W/A} & \multicolumn{2}{c}{\multirow{2}{*}{Market-1501}} & \multicolumn{2}{c}{\multirow{2}{*}{\begin{tabular}[c]{@{}c@{}}DukeMTMC\\ -reID\end{tabular}}} & \multicolumn{2}{c}{\multirow{2}{*}{CUHK03}} & \multirow{3}{*}{{\begin{tabular}[c]{@{}c@{}}Search\\Method\end{tabular}}} \\
                              &                         &                      & \multicolumn{2}{c}{}                             & \multicolumn{2}{c}{}                                                                          & \multicolumn{2}{c}{}                        &                                                                          \\ \cline{4-9}
                              &                         & (bit)                & mAP                     & R1                     & mAP                                            & R1                                           & mAP                   & R1                  &                                                                          \\ \hline
        \multicolumn{1}{c}{\multirow{4}{*}{\re{ResNet-18}}}                     & Real-valued                    & 32/32                                                                & 64.3                                               & 85.1                                                  & 48.1                                               & 72.9                                                  & 33.0                                               & 42.7                                                  & \multirow{4}{*}{Manual}                                                       \\ \cline{2-9}
        \multicolumn{1}{c}{}                                           & XNOR-Net                       & \multirow{3}{*}{1/1}                                                 & 40.1                                               & 63.8                                                  & 24.9                                               & 52.1                                                  & 14.0                                               & 26.1                                                  &                                                                               \\
        \multicolumn{1}{c}{}                                           & Bi-Real Net                    &                                                                      & 52.7                                               & 73.1                                                  & 36.3                                               & 64.9                                                  & 25.6                                               & 36.0                                                  &                                                                               \\
        \multicolumn{1}{c}{}                                           & BiRe-ID                    &                                                                      & 64.0                                               & 84.1                                         & 47.1                                               & 70.7                                                  & 31.8                                               & 40.9                                                  &                                                                               \\ \hline
        \multicolumn{2}{c}{{\begin{tabular}[c]{@{}c@{}}\textbf{DCP-NAS-S}\end{tabular}}} & {1/1}                                              & {\textbf{68.0}}                     & {\textbf{84.9}}                                 & {\textbf{60.2}}                     & {\textbf{76.4}}                        & {\textbf{48.8}}                     & {\textbf{50.0}}                        & Gradient-based$^\natural$ \\ \hline
        \multicolumn{1}{c}{\multirow{4}{*}{\re{ResNet-34}}}                     & Real-valued                    & 32/32                                                                & 68.0                                               & 86.7                                                  & 52.2                                               & 73.9                                                  & 39.2                                               & 47.9                                                  & \multirow{4}{*}{Manual}                                                                        \\ \cline{2-9}
        \multicolumn{1}{c}{}                                           & XNOR-Net                       & \multirow{3}{*}{1/1}                                                 & 45.9                                               & 67.1                                                  & 27.7                                               & 55.7                                                  & 18.4                                               & 29.6                                                  &                                                                               \\
        \multicolumn{1}{c}{}                                           & Bi-Real Net                    &                                                                      & 54.1                                               & 77.6                                                  & 40.8                                               & 69.1                                                  & 29.3                                               & 38.2                                                  &                                                                               \\
        \multicolumn{1}{c}{}                                           & BiRe-ID                    &                                                                      & 67.1                                               & 85.3                                                  & 51.0                                               & 72.4                                                  & 37.1                                               & 46.4                                                  &                                                                               \\ \hline
        \multicolumn{2}{c}{\begin{tabular}[c]{@{}c@{}}\textbf{DCP-NAS-M} \end{tabular}}                  & 1/1                                                                  & \textbf{70.6}                                      & \textbf{86.2}                                         & \textbf{61.9}                                      & \textbf{77.9}                                         & \textbf{51.4}                                      & \textbf{53.6}                                         & Gradient-based$^\natural$ \\ \hline
        \multicolumn{1}{c}{\multirow{4}{*}{\re{ResNet-50}}}                     & Real-valued                    & 32/32                                                                & 71.6                                               & 88.8                                                  & 56.9                                               & 75.4                                                  & 43.8                                               & 50.6                                                  & \multirow{4}{*}{Manual}                                                       \\ \cline{2-9}
        \multicolumn{1}{c}{}                                           & XNOR-Net                       & \multirow{3}{*}{1/1}                                                 & 49.1                                               & 68.3                                                  & 31.1                                               & 59.4                                                  & 20.3                                               & 31.5                                                  &                                                                               \\
        \multicolumn{1}{c}{}                                           & Bi-Real Net                    &                                                                      & 58.9                                               & 78.5                                                  & 45.1                                               & 71.0                                                  & 34.7                                               & 42.6                                                  &                                                                               \\
        \multicolumn{1}{c}{}                                           & BiRe-ID                    &                                                                      & 69.2                                               & 86.9                                                  & 55.0                                               & 74.1                                                  & 41.1                                               & 49.0                                                  &                                                                               \\ \hline
       \multicolumn{2}{c}{\begin{tabular}[c]{@{}c@{}}\textbf{DCP-NAS-L} \end{tabular}}                   & 1/1                                                                  & \textbf{71.9}                                      & \textbf{87.3}                                         & \textbf{62.4}                                      & \textbf{78.7}                                         & \textbf{54.4}                                      & \textbf{57.6}                                         & Gradient-based$^\natural$                \\ \bottomrule
        \end{tabular}}
        \vspace{-4mm}
        \label{reid}
    \end{table}

    \subsection{Results on Person Re-identification}
    \label{sec:identification} 
    We implement our experiments on three mainstream datasets of person Re-ID task, {\em i.e.}, Market1501, DukeMTMC-reID, and CUHK03. We report the performance of real-valued models, previous binarization methods XNOR-Net~\cite{bulat2019xnor} and Bi-Real Net~\cite{liu2018bi}, for comparison. We also compare our DCP-NAS with state-of-the-art binarization person reidentification work BiRe-ID~\cite{xu2022bire}. Three mainstream backbones are chosen for comparison: ResNet-18, ResNet-34, and ResNet-50.
    
    \noindent{\bf Market1501:} As shown in the column 3 of Tab.~\ref{reid}, on ResNet-18, ResNet-34 and ResNet-50 backbones, our DCP-NAS-S, DCP-NAS-M and DCP-NAS-L both surpass the real-valued counterpart in a clear margin (68.0 $\%$ mAP vs. 64.3 $\%$ mAP, 70.6 $\%$ mAP vs. 68.0 $\%$ mAP, 71.9 $\%$ mAP vs. 71.6 $\%$ mAP). Our DCP-NAS outperforms other 1-bit models (XNOR-Net and Bi-Real Net) by a sizable margin. Moreover, our DCP-NAS outperforms BiRe-ID by 4.0\%, 3.5\%, 2.7\% mAP with different model sizes, respectively. 
    
    \noindent{\bf DukeMTMC-reID:} On DukeMTMC-reID dataset, our DCP-NAS also achieves significant performances, as listed in the column 4 of Tab.~\ref{reid}. On ResNet-18, ResNet-34 and ResNet-50 backbones, our DCP-NAS-S, DCP-NAS-M and DCP-NAS-L both outperforms the real-valued models by 12.1\%, 9.7\%, 5.5\% mAP. Our DCP-NAS outperforms other 1-bit models (XNOR-Net and Bi-Real Net) by a sizable margin. Moreover, our DCP-NAS outperforms BiRe-ID by 13.1\%, 10.9\%, 7.4\% mAP with different model sizes, respectively. 
    
    \noindent{\bf CUHK03:} As shown in the column 5 of Tab.~\ref{reid}, our DCP-NAS shows significantly representation ability on person ReID task in the CUHK03 dataset. With ResNet-18, ResNet-34 and ResNet-50 backbones, our DCP-NAS-S, DCP-NAS-M and DCP-NAS-L both achieve clear performance advantage than the real-valued models (48.8 $\%$ mAP vs. 33.0 $\%$ mAP, 51.4 $\%$ mAP vs. 39.2 $\%$ mAP, 54.4 $\%$ mAP vs. 43.8 $\%$ mAP). Our DCP-NAS surpasses other 1-bit models (XNOR-Net and Bi-Real Net) by a sizable margin. Moreover, our DCP-NAS outperforms BiRe-ID by 15.8\%, 12.2\%, 10.6\% mAP with different model sizes, respectively.

    \begin{table*}[]
        \caption{Comparison of mAP$(\%)$ for objects with state-of-the-art \re{1-bit} object detectors in  Faster-RCNN detection framework on VOC {\tt test2007}, where the performances of real-valued and 4-bit detectors are reported for reference. Gradient-based$^\natural$ denotes our DCP-NAS is a discrepancy-aware gradient-based search method.}
        \renewcommand\arraystretch{1.3}
        \centering
        \begin{tabular}{ccccccc}
        \hline
        Backbone & Quantization                          & W/A(bit)             & \begin{tabular}[c]{@{}c@{}}mAP\\ (\%)\end{tabular} & \begin{tabular}[c]{@{}c@{}}Mem.\\ (MB)\end{tabular} & \begin{tabular}[c]{@{}c@{}}OPs\\ ($\times 10^9$)\end{tabular} & \begin{tabular}[c]{@{}c@{}}Search\\ Method\end{tabular}      \\ \hline
        \multicolumn{1}{c}{\multirow{4}{*}{ResNet-18}} & Real-valued & 32/32                & 76.4                                               & 112.88                                              & 96.40                                                         & \multirow{4}{*}{Manual}                                      \\ \cline{2-6}
        \multicolumn{1}{c}{}                           & DoReFa-Net  & 4/4                  & 73.3                                               & 21.59                                               & 27.15                                                         &                                                              \\\cline{2-6}
        \multicolumn{1}{c}{}                           & ReActNet    & \multirow{2}{*}{1/1} & 69.6                                               & \multirow{2}{*}{16.61}                              & \multirow{2}{*}{18.49}                                        &                                                              \\
        \multicolumn{1}{c}{}                           & LWS-Det     &                      & 73.2                                               &                                                     &                                                               &                                                              \\ \hline
        \multicolumn{2}{c}{\textbf{DCP-NAS-S}}                          & 1/1                  & \textbf{72.8}                                              & 13.42                                               & 15.01                                                         & Gradient-based$^\natural$ \\ \hline
        \multicolumn{2}{c}{\textbf{DCP-NAS-M}}                         & 1/1                  & \textbf{74.2}                                      & 13.56                                               & 15.10                                                         & Gradient-based$^\natural$\\ \hline
        \multicolumn{1}{c}{\multirow{4}{*}{ResNet-34}} & Real-valued & 32/32                & 77.8                                               & 145.12                                              & 118.80                                                        & \multirow{4}{*}{Manual}                                      \\ \cline{2-6}
        \multicolumn{1}{c}{}                           & DoReFa-Net  & 4/4                  & 75.6                                               & 29.65                                               & 32.31                                                         &                                                              \\\cline{2-6}
        \multicolumn{1}{c}{}                           & ReActNet    & \multirow{2}{*}{1/1} & 72.3                                               & \multirow{2}{*}{24.68}                              & \multirow{2}{*}{21.49}                                        &                                                              \\
        \multicolumn{1}{c}{}                           & LWS-Det     &                      & 75.8                                               &                                                     &                                                               &                                                              \\ \hline
        \multicolumn{2}{c}{\textbf{DCP-NAS-L}}                          & 1/1                  & \textbf{76.7}                                      &  13.79                                              & 15.29                                                         & Gradient-based$^\natural$ \\ \hline
        \end{tabular}
        \label{VOC}
    \end{table*}
    
    \subsection{{Results on Object Detection}}\label{sec:detection}
    \subsubsection{{Results on the PASCAL VOC dataset}}
    In this section, we compare the proposed DCP-NAS with other state-of-the-art 1-bit detectors based on Faster-RCNN, including ReActNet~\cite{liu2020reactnet}, and LWS-Det~\cite{xu2021layer}, on the same framework for the task of object detection on the PASCAL VOC datasets. We also report the detection performance of the multi-bit quantized network DoReFa-Net~\cite{zhou2016dorefa}.
    Tab. \ref{VOC} illustrates the comparison of the mAP across different quantization methods and detection frameworks. Our DCP-NAS efficiently generalizes to the downstream task and significantly accelerates the computation and saves the storage on various detectors by binarizing the activations and weights to 1-bit.
    The results for 1-bit Faster-RCNN on VOC {\tt test2007} are summarized in Tab. \ref{VOC}. Compared with other 1-bit methods, we observe significant performance improvements with our DCP-NAS over other state-of-the-arts. Compared with the ResNet-18 backbone, our DCP-NAS-S outperforms ReActNet $3.2\%$ mAP with the same bit-width activations and weights, but higher acceleration rate (15.01 GOPs vs. 18.49 GOPs). Our DCP-NAS-M achieves surpasses other low-bit detectors including DoReFa-Net, ReActNet and LWS-Det by $0.9\%$, $4.6\%$ and $1.0\%$.
    The lines 8 to 12 in Tab. \ref{VOC} illustrate that our DCP-NAS-L achieves a performance far more close to the real-valued counterpart compared with other prior low-bit detectors with ResNet-34 backbone. Quantitatively speaking, our DCP-NAS surpasses DoReFa-Net, ReActNet and LWS-Det by $1.1\%$, $4.4\%$, and $0.9\%$ mAP, respectively. Moreover, DCP-NAS even achieves comparable performance as real-valued ($76.7\%$ vs. $77.8\%$) with high compression rate and speed acceleration, which demonstrates the superiority of our DCP-NAS.
    
    In short, we achieved a new state-of-the-art performance compared to other 1-bit CNNs on various detection frameworks with various backbones on PASCAL VOC. We also surpass the performance to \re{real-valued} models, as demonstrated in extensive experiments, clearly validating the superiority of our DCP-NAS.

    \begin{table}[]
        \small
        \centering
        \caption{Comparison of mAP$@[.5, .95] (\%)$, AP with different IoU threshold and AP for objects in various sizes with state-of-the-art \re{1-bit} object detectors in  Faster-RCNN framework on COCO {\tt minival}, where the performances of real-valued and 4-bit detectors are reported for reference. Gradient-based$^\natural$ denotes our DCP-NAS is a discrepancy-aware gradient-based search method.}
        \renewcommand\arraystretch{1}
        \setlength{\tabcolsep}{0.5mm}{
        \begin{tabular}{cccccccccc}
        \toprule
        Backbone & Quantization                           & W/A(bit)             & \begin{tabular}[c]{@{}c@{}}mAP\\ @{[}.5, .95{]}\end{tabular} & AP$_{50}$     & AP$_{75}$     & AP$_{s}$      & AP$_{m}$      & AP$_{l}$      & \begin{tabular}[c]{@{}c@{}}Search\\ Method\end{tabular}      \\ \hline
        \multicolumn{1}{c}{\multirow{4}{*}{ResNet-18}} & Real-valued & 32/32                &32.2 &53.8 &34.0 &18.0 &34.7 &41.9          & \multirow{4}{*}{Manual}                                      \\ \cline{2-9}
        \multicolumn{1}{c}{}                           & FQN  & 4/4                  &28.1 &48.4 &29.3 &14.5 &30.4 &38.1&                                                              \\\cline{2-9}
        \multicolumn{1}{c}{}                           & ReActNet &    \multirow{2}{*}{1/1}                 &21.1 &38.5 &20.5 &9.7 &23.5 &32.1&                                                              \\
        \multicolumn{1}{c}{}                           & LWS-Det       &                &26.9 &44.9 &27.7 &12.9 &28.7 &38.3&                                                              \\ \hline
        \multicolumn{2}{c}{\textbf{DCP-NAS-S}}                          & {1/1} & \textbf{25.4}                                                & \textbf{42.9} & \textbf{24.6} & \textbf{14.6} & \textbf{26.3} & \textbf{33.8} & Gradient-based$^\natural$ \\ \hline
        \multicolumn{2}{c}{\textbf{DCP-NAS-M}}                         & {1/1} & \textbf{27.3}                                                & \textbf{45.2} & \textbf{27.4} & \textbf{17.4} & \textbf{28.5} & \textbf{35.6} & Gradient-based$^\natural$ \\ \hline
        \multicolumn{2}{c}{\textbf{DCP-NAS-L}}                          & {1/1} & \textbf{29.7}                                                & \textbf{47.8} & \textbf{29.9} & \textbf{19.5} & \textbf{30.7} & \textbf{37.5} & Gradient-based$^\natural$ \\ \bottomrule
        \end{tabular}}
        \label{COCO}
    \end{table}

    \subsubsection{{Results on the COCO dataset}}
    The COCO dataset is much more challenging for object detection than PASCAL VOC due to its diversity and scale. We compare the proposed DCP-NAS with state-of-the-art \re{1-bit CNNs}, including XNOR-Net~\cite{rastegari2016xnor}, Bi-Real-Net~\cite{liu2020bi}, and BiDet~\cite{wang2020bidet}, on COCO. We also report the detection performance of the 4-bit quantized DoReFa-Net~\cite{zhou2016dorefa}.
    Tab. \ref{COCO} shows  mAP and AP with different IoU thresholds and  AP of objects with different scales. Limited by the page width, we do not show the memory usage and FLOPs in Tab. \ref{COCO}. We conduct experiments on Faster-RCNN framework.
    Compared with the state-of-the-art 1-bit methods, our DCP-NAS outperforms other methods by significant margins. Our DCP-NAS-S improves the mAP$@[.5,.95]$ by $4.3\%$ compared with state-of-the-art ReActNet and our DCP-NAS-M surpasses the state-of-the-art 1-bit detectors ReActNet and LWS-Det by $6.2\%$ and $0.4\%$ on mAP$@[.5,.95]$, respectively. Compared with the 4-bits detectors FQN, our DCP-NAS-L improves the mAP$@[.5,.95]$ by $1.6\%$ with binary weights and activations which can significantly compress the model parameters and accelerate the inference process. Our DCP-NAS-L also achieves much closer performance on mAP$@[.5,.95]$ compared with real-valued ResNet-18 based Faster-RCNN ($29.7\%$ vs. $32.2\%$), but with extremely compressed detectors' weights and activations. Similarly, on other APs with different IoU thresholds, our DCP-NAS outperforms other methods obviously. 
    
    To conclude, compared with the baseline methods of network quantization, our method achieves the best performance in terms of the AP with different IoU thresholds and the AP for objects in different sizes on COCO, demonstrating DCP-NAS's superiority and universality in different application settings.
    
    \subsection{Deployment Efficiency}
    \label{deploy}

    \begin{table}[t]
        \centering
        \caption{Comparing DCP-NAS-M with real-valued ResNet-18 and \re{1-bit} ResNet-18 (ReActNet) on hardware (single thread).}
        \renewcommand\arraystretch{1.05}
        \begin{tabular}{cccccc}
        \toprule
        \multirow{2}{*}{Model}                                         & \multirow{2}{*}{W/A} & Memory  & Memory       & Latency & \multirow{2}{*}{Acceleration} \\
                                                                   &                      & (MBits) & Saving       & (ms)    &                               \\ \hline
        ResNet-18                                                      & 32/32                & 357.4   & -            & 255.7  & -                             \\ \hline
        \begin{tabular}[c]{@{}c@{}}ResNet-18\\ (ReActNet)\end{tabular} & 1/1                  & 11.2    & 31.9$\times$ & 67.1    & 3.8$\times$                  \\ \hline
        DCP-NAS-M                                                      & 1/1                  & 6.9     & 51.8$\times$ & 101.3    & 2.5$\times$                  \\ \bottomrule
        \end{tabular}
        \label{hardware}
    \end{table}

    We implement the 1-bit models achieved by our DCP-NAS-M on ODROID C4, which has a 2.016 GHz 64-bit quad-core ARM Cortex-A55. With evaluating its real speed in practice, the efficiency of our DCP-NAS is proved when deployed into real-world mobile devices. 
    We leverage the SIMD instruction SSHL on ARM NEON to make the inference framework BOLT~\cite{feng2021bolt} compatible with DCP-NAS.
    We compare DCP-NAS-M to the real-valued ResNet-18 and bianrized ResNet-18 (ReActNet) in Tab. \ref{hardware}. We can see that DCP-NAS-M's inference speed is substantially faster with the highly efficient BOLT framework. For example, the acceleration rate achieves about 2.5$\times$ compared with ResNet-18, which is slightly lower than the acceleration rate of ReActNet's 3.8$\times$. However, our DCP-NAS-M has higher memory saving ratio, compared with ReActNet. The deployment efficiency experiment reveals the effectiveness and efficiency of our DCP-NAS. All deployment results are significant for the computer vision on real-world edge devices.

    \section{Conclusion and future work}
    This paper proposes Discrepant Child-Parent Neural Architecture search (DCP-NAS), which consider the architecture discrepancy between real-valued and bianrized models, resulting in a unified binary neural architecture search framework with novel decoupled optimization. We incorporate tangent propagation, GGN method and decoupled backtracking optimization into the binary NAS in a efficient and effective manner. 

    Extensive experiments on CIFAR and ImageNet demonstrate that DCP-NAS achieves the best classification performance comparing with both previous binary NAS and directly binarizing real-valued NAS, and even surpasses real-valued ResNet-18 in a clear margin. 
    We also achieve a promising performance on person re-identification and object detection, which validate the generality of our proposed method. In the future, we will extend our method with Transformer-based architectures to build more generalized binary NAS. We will also try other optimization methods to find the optimal architecture for more compact neural networks.

    \section{Acknowledgement}
    This work was supported in part by the National Natural Science Foundation of China under Grant 62076016 and 61827901, Beijing Natural Science Foundation-Xiaomi Innovation Joint Fund L223024, Foundation of China Energy Project GJNY-19-90. “One Thousand Plan” innovation leading talent funding projects in Jiangxi Province Jxsg2023102268

\bibliographystyle{spmpsci}      
\bibliography{bibliography}   

%
%

\end{document}